\documentclass[10pt,twocolumn,letterpaper]{article}

\usepackage{cvpr}              

\usepackage{graphicx}
\usepackage{amsmath}
\usepackage{amssymb}
\usepackage{booktabs}

\usepackage[pagebackref,breaklinks,colorlinks]{hyperref}

\usepackage[capitalize]{cleveref}
\crefname{section}{Sec.}{Secs.}
\Crefname{section}{Section}{Sections}
\Crefname{table}{Table}{Tables}
\crefname{table}{Tab.}{Tabs.}

\def\cvprPaperID{10285} 
\def\confName{CVPR}
\def\confYear{2022}

\newsavebox\CBox
\def\textBF#1{\sbox\CBox{#1}\resizebox{\wd\CBox}{\ht\CBox}{\textbf{#1}}}

\begin{document}

\title{MicroISP: Processing 32MP Photos on Mobile Devices with Deep Learning}

\author{Andrey Ignatov$^{1,2}$ \and Anastasia Sycheva$^1$ \and Radu Timofte$^{1,2}$ \and Yu Tseng$^3$ \and Yu-Syuan Xu$^3$ \and Po-Hsiang Yu$^3$ \and Cheng-Ming Chiang$^3$ \and Hsien-Kai Kuo$^3$ \and Min-Hung Chen$^3$ \and Chia-Ming Cheng$^3$ \hspace{36mm} Luc Van Gool$^{1,2}$ \vspace{3.2mm}\\
$^{1}$ ETH Zurich, Switzerland \hspace{6mm} $^{2}$ AI Witchlabs Ltd., Switzerland \hspace{6mm} $^{3}$ MediaTek Inc., Taiwan}

\maketitle

\begin{abstract}
   While neural networks-based photo processing solutions can provide a better image quality compared to the traditional ISP systems, their application to mobile devices is still very limited due to their very high computational complexity. In this paper, we present a novel MicroISP model designed specifically for edge devices, taking into account their computational and memory limitations. The proposed solution is capable of processing up to 32MP photos on recent smartphones using the standard mobile ML libraries and requiring less than 1 second to perform the inference, while for FullHD images it achieves real-time performance. The architecture of the model is flexible, allowing to adjust its complexity to devices of different computational power. To evaluate the performance of the model, we collected a novel Fujifilm UltraISP dataset consisting of thousands of paired photos captured with a normal mobile camera sensor and a professional 102MP medium-format FujiFilm GFX100 camera. The experiments demonstrated that, despite its compact size, the MicroISP model is able to provide comparable or better visual results than the traditional mobile ISP systems, while outperforming the previously proposed efficient deep learning based solutions. Finally, this model is also compatible with the latest mobile AI accelerators, achieving good runtime and low power consumption on smartphone NPUs and APUs. The code, dataset and pre-trained models are available on the project website: \url{https://people.ee.ethz.ch/~ihnatova/microisp.html}

\end{abstract}

\vspace{-3.2mm}
\section{Introduction}
\label{sec:intro}

\begin{figure*}[t!]
\centering
\setlength{\tabcolsep}{1pt}
\resizebox{\linewidth}{!}
{
\begin{tabular}{ccc}
\scriptsize{Visualized RAW Image}\normalsize & \scriptsize{MediaTek Dimensity 820 ISP Photo}\normalsize & \scriptsize{Fujifilm GFX 100 Photo}\normalsize\\
    \includegraphics[width=0.33\linewidth]{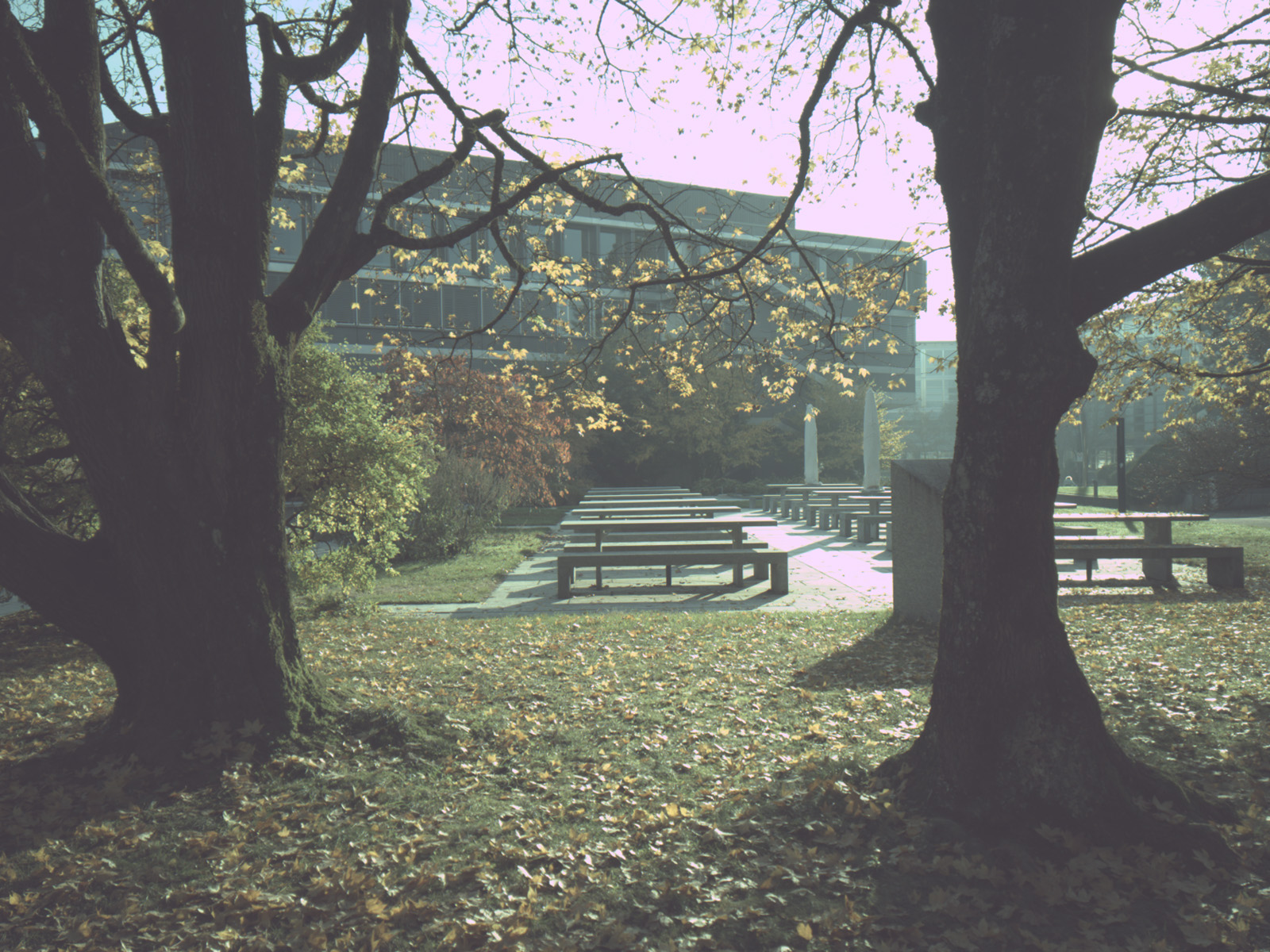}&
    \includegraphics[width=0.33\linewidth]{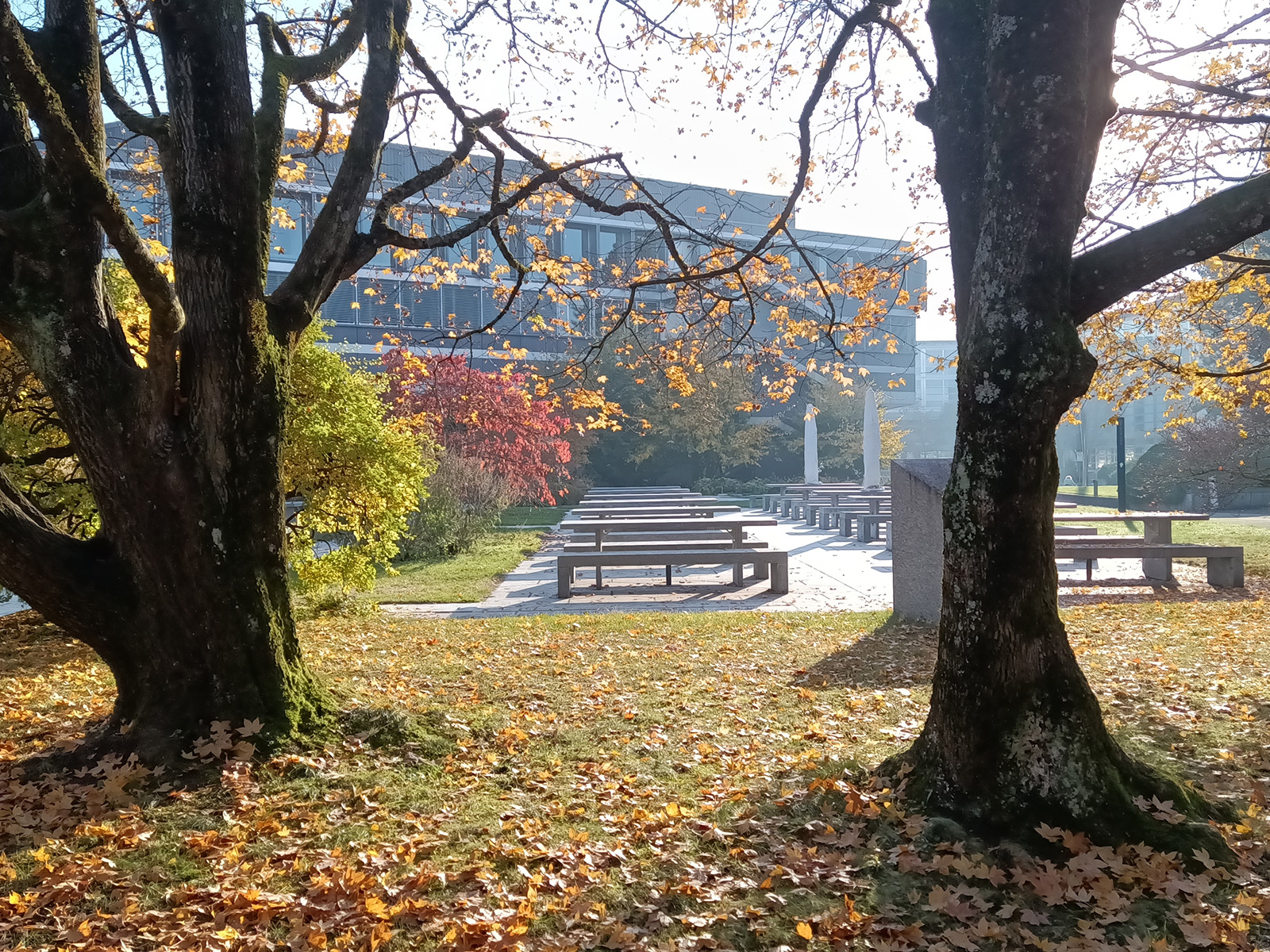}&
    \includegraphics[width=0.33\linewidth]{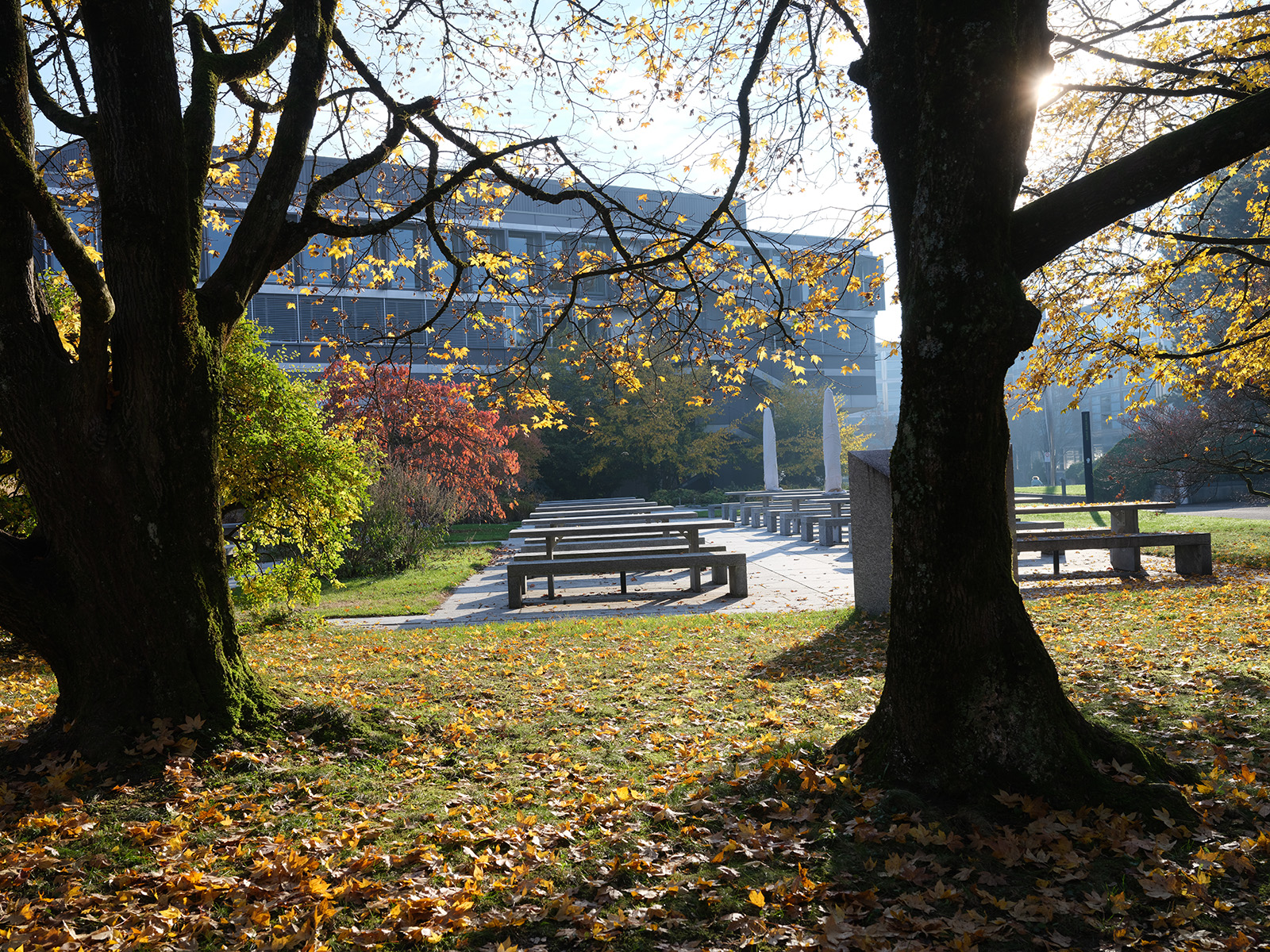} \\
    \includegraphics[width=0.33\linewidth]{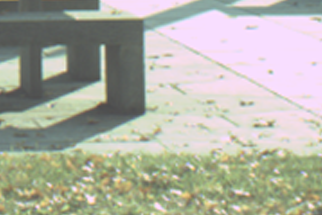}&
    \includegraphics[width=0.33\linewidth]{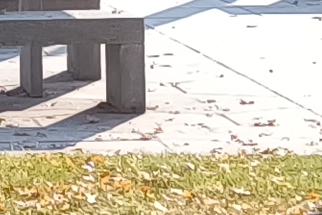}&
    \includegraphics[width=0.33\linewidth]{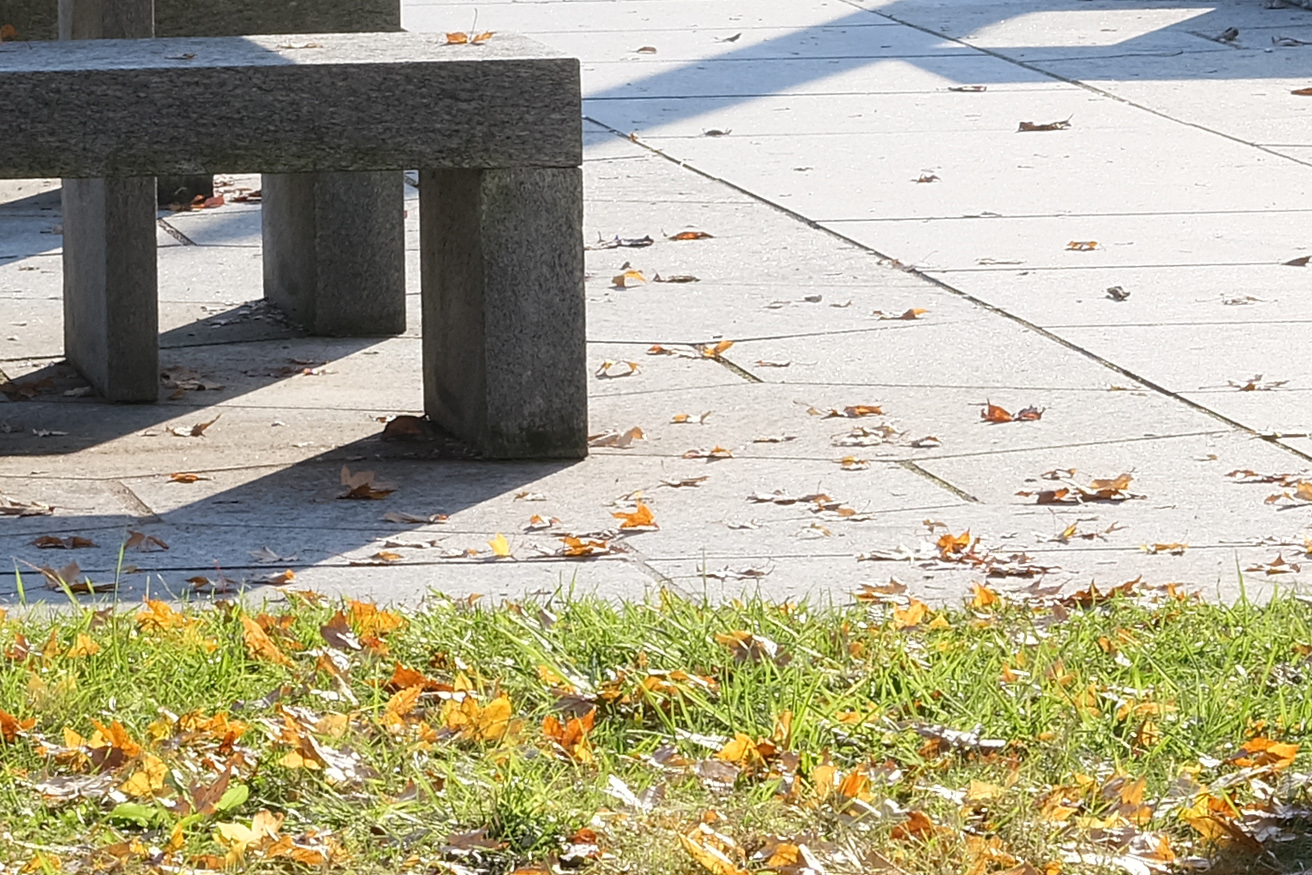} \\
\end{tabular}
}
\vspace{-0.2cm}
\caption{Example set of full-resolution images (top) and crops (bottom) from the collected Fujifilm UltraISP dataset. From left to right: original RAW visualized image, RGB image obtained with MediaTek's built-in ISP system, and Fujifilm GFX100 target photo.}
\vspace{-0.2cm}
\label{fig:dataset}
\end{figure*}

As camera quality becomes a prime feature of smartphones, the requirements on mobile photos grow each year. Developments in this field follow two directions: enhancements in camera sensor and optics hardware, and improvements related to computational photography. As the hardware limitations on sensor size and resolution are almost reached, the latter option plays an even more important role.

The problem of image restoration and enhancement has been addressed in several papers, though many were dealing only with particular aspects such as super-resolution~\cite{dong2015image,kim2016accurate,lim2017enhanced,timofte2017ntire,timofte2018ntire,cai2019ntire,lugmayr2020ntire,zhang2020ntire,ignatov2018pirm,ignatov2021real}, denoising~\cite{gu2019brief,tai2017memnet,zhang2018ffdnet,zhang2017beyond,abdelhamed2020ntire,abdelhamed2019ntire,ignatov2021fast}, color and tone mapping~\cite{salih2012tone,ma2014high,yan2016automatic,lee2016automatic}, luminance, gamma and contrast adjustment~\cite{yuan2012automatic,fu2016fusion,cai2018learning}. Comprehensive end-to-end photo quality enhancement has first been addressed in~\cite{ignatov2018pirm,ignatov2019ntire}, where the authors proposed to learn a mapping between low-quality RGB smartphone images and target high-quality DSLR photos with deep learning. Despite huge subsequent progress~\cite{vu2018fast,lugmayr2019unsupervised,de2018fast,hui2018perception,huang2018range,liu2018deep,ignatov2018pirm,ignatov2019ntire}, this approach had a significant limitation: the photos that one gets with smartphone ISPs undergo many image processing steps that heavily alter the original pixel data, and lead to a severe information loss caused by noise suppression, narrowing of the original dynamic range, jpeg compression, etc. Thus, one can get much better results when working directly with the original RAW sensor data. This approach was explored in~\cite{ignatov2020replacing,ignatov2020aim,dai2020awnet,silva2020deep,kim2020pynet,ignatov2019aim}, where the authors presented the Zurich RAW-to-RGB dataset and obtained results comparable or better than the ones of the ISP system of the Huawei P20 smartphone. This was an important proof of concept showing it is possible to replace conventional hand-crafted ISP pipelines with end-to-end deep learning approaches, though a significant limitation was left: the models were too heavy to run on real mobile devices. Thus, the practical application of these solutions was very limited. An important step in solving this problem was done in the Mobile AI Challenge~\cite{ignatov2021learned}, where the participants were developing efficient ISP models for inference on mobile devices. However, the size of the photos in this challenge was still limited to 2MP, while the resolution of real smartphone cameras is at least 12MP and can be as high as 108MP.

In this paper, we propose the first deep learning based solution able to process 32MP RAW photos on smartphones and designed taking into account their hardware limitations. As the current public RAW-to-RGB dataset~\cite{ignatov2020replacing} has issues with the quality of the target images and their alignment to the original RAW data, we collect a novel large-scale dataset using a professional medium format camera capturing 102MP photos and a Sony mobile sensor for getting the original RAW data, and align pixel-wise the obtained photos. Finally, we present experiments evaluating the quality of the resulting RGB images and the runtime of the solution on recent flagship mobile platforms and AI accelerators.

The remainder of the paper is structured as follows. In Section~\ref{sec:dataset} we describe the new Fujifilm UltraISP dataset. Section~\ref{sec:architecture} presents our MicroISP architecture and describes the underlying design choices. Section~\ref{sec:experiments} shows
and analyzes the experimental results and discusses the limitations of the solution. Finally, Section~\ref{sec:conclusion} concludes the paper.

\section{Fujifilm UltraISP Dataset}
\label{sec:dataset}

When dealing with an end-to-end learned smartphone ISP, the quality of the target images used for training the model plays a crucial. Thus, the requirements on the target camera are high: it should produce photos that are outstanding in terms of real resolution, noise free even when captured in low light conditions, exhibit a high dynamic range and pleasant color rendition, and are sharp enough when shooting them with an open aperture. As our exploration revealed that none of the currently existing APS-C and full-frame cameras satisfy all those requirements, we used the Fujifilm GFX100, a medium format 102 MP camera, for capturing the target high-quality photos. To collect the source RAW smartphone images, we chose a popular Sony IMX586 Quad Bayer camera sensor that can be found in tens of mid-range and high-end mobile devices released in the past 3 years. This sensor was mounted on the \mbox{MediaTek} Dimensity 820 development board, and was capturing both raw and processed (by its built-in ISP system) 12MP images. The Dimensity board was rigidly attached to the Fujifilm camera and controlled using a specialized software developed for this project. The cameras were capturing photos synchronously to ensure that the image content is identical. This setup was used for several weeks to collect over 6 thousand daytime image pairs at a wide variety of places with different illumination and weather conditions. An example set of full-resolution photos from the collected dataset is shown in Fig.~\ref{fig:dataset}.

As the collected RAW-RGB image pairs were not perfectly aligned, we had to perform local matching first. In order to achieve a precise pixel-wise alignment, we used the SOTA deep learning based dense matching algorithm~\cite{truong2021learning} to extract 256$\times$256 px patches from the original photos. This procedure resulted in over 99K pairs of crops that were divided into training (93.8K), validation (2.2K) and test (3.1K) sets and used for model training and evaluation. It should be mentioned that all alignment operations were performed on Fujifilm RGB images only, therefore RAW photos from the Sony sensor remained unmodified, exhibiting exactly the same values as read from the sensor.

\section{Architecture}
\label{sec:architecture}

\begin{figure*}[t!]
\centering
\resizebox{\linewidth}{!}{
\includegraphics[width=1.0\linewidth]{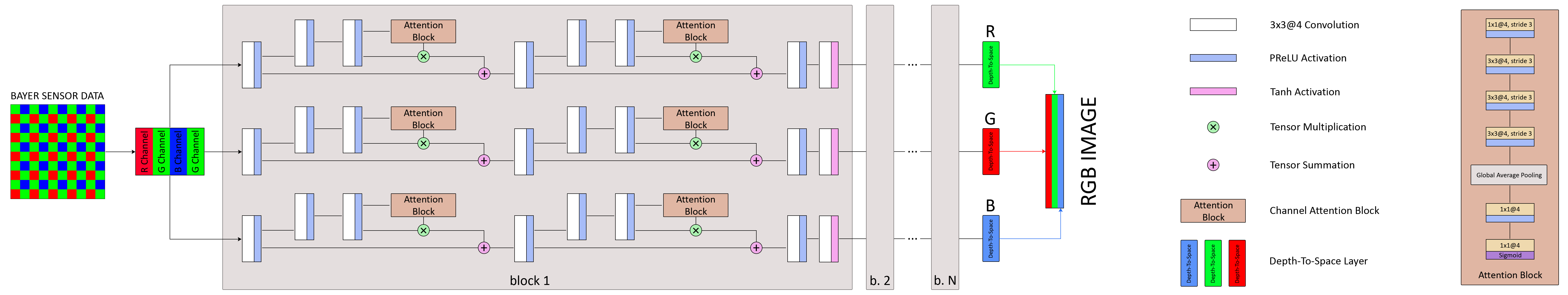}
}
\vspace{-0.2cm}
\caption{The overall architecture of the proposed MicroISP model (left), and the structure of the enhanced attention block (right).}
\vspace{-0.2cm}
\label{fig:architecture}
\end{figure*}

When designing a model capable of processing high-resolution images on mobile devices, one needs to address the following limitations related to edge inference:

\begin{itemize}
\item \textBF{Memory consumption:} unlike the standard desktop systems, mobile devices have a limited amount of accessible RAM. This restriction becomes even more severe when the inference happens on mobile AI accelerators such as NPUs or APUs that usually have their own memory limited to hundreds of megabytes.
\item \textBF{Layers and operators:} mobile AI accelerators support only a restricted set of common machine learning ops. Thus, on the latest Android devices, one is limited to 101 different operators at maximum~\cite{NNAPIDrivers2021,NNAPI13Specs}, while older mobile AI accelerators might be supporting even less than 28 layers~\cite{NNAPI10Specs} (including the basic ones such as summation, multiplication, convolution, \etc).
\item \textBF{Computational complexity:} with common architectures such as U-Nets / ResNets / PyNETs, it takes tens of seconds to process 32MP images on desktop or server systems with high-end Nvidia GPUs. As even the latest smartphone AI accelerators are considerably less powerful, model complexity should be very low in order to achieve a runtime of 1-2 secs.
\item \textBF{Model size:} since the resulting NN model is usually integrated in the camera application, its size should be reasonably small, not exceeding several megabytes.
\end{itemize}

\begin{figure*}[t!]
\centering
\setlength{\tabcolsep}{1pt}
\resizebox{\linewidth}{!}
{
\begin{tabular}{cccc}
\scriptsize{Visualized RAW Image}\normalsize & \scriptsize{MediaTek Dimensity 820 ISP Photo}\normalsize & \scriptsize{Reconstructed RGB Image (MicroISP)}\normalsize & \scriptsize{Fujifilm GFX 100 Photo}\normalsize\\
    \includegraphics[width=0.24\linewidth]{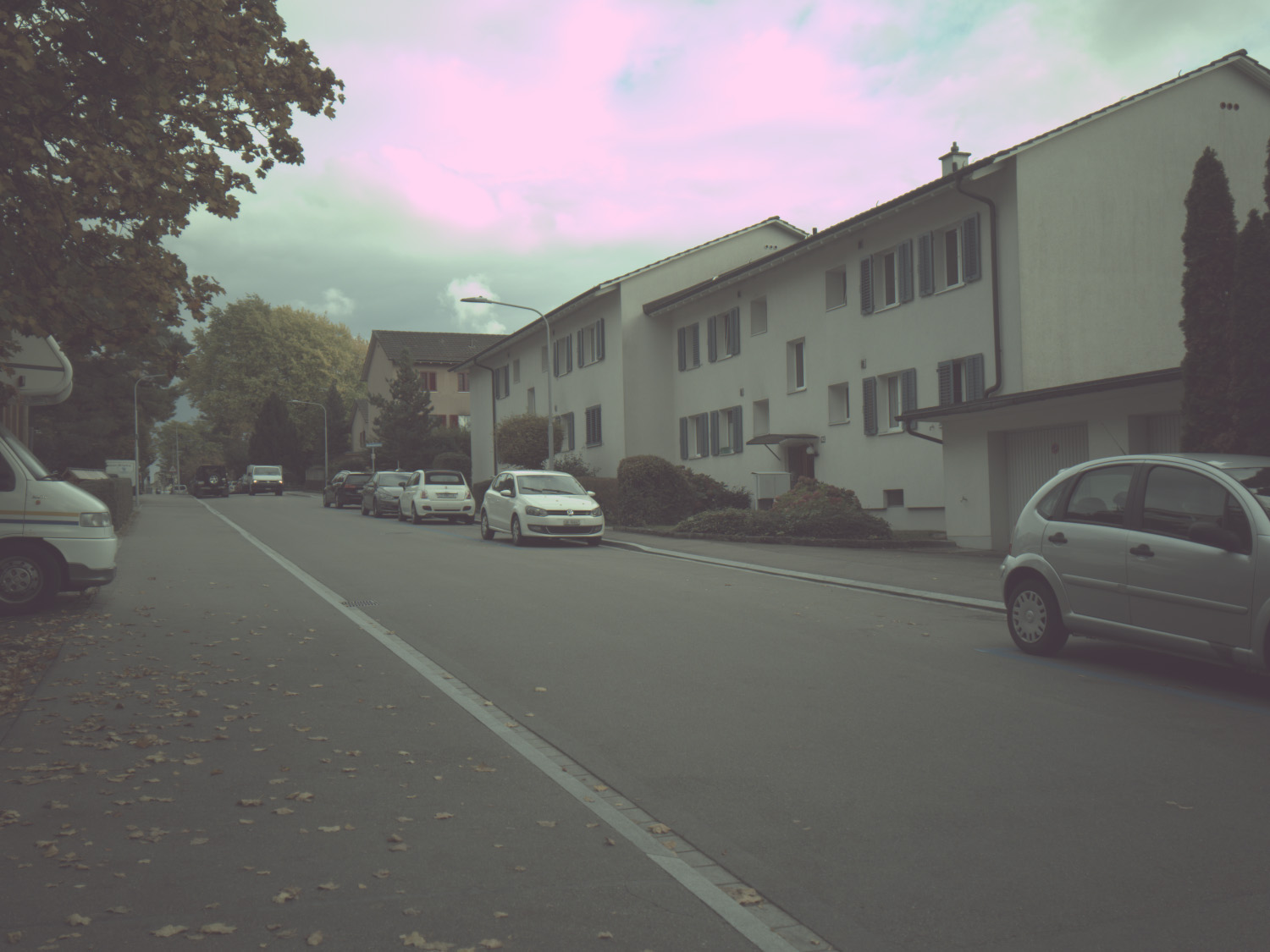}&
    \includegraphics[width=0.24\linewidth]{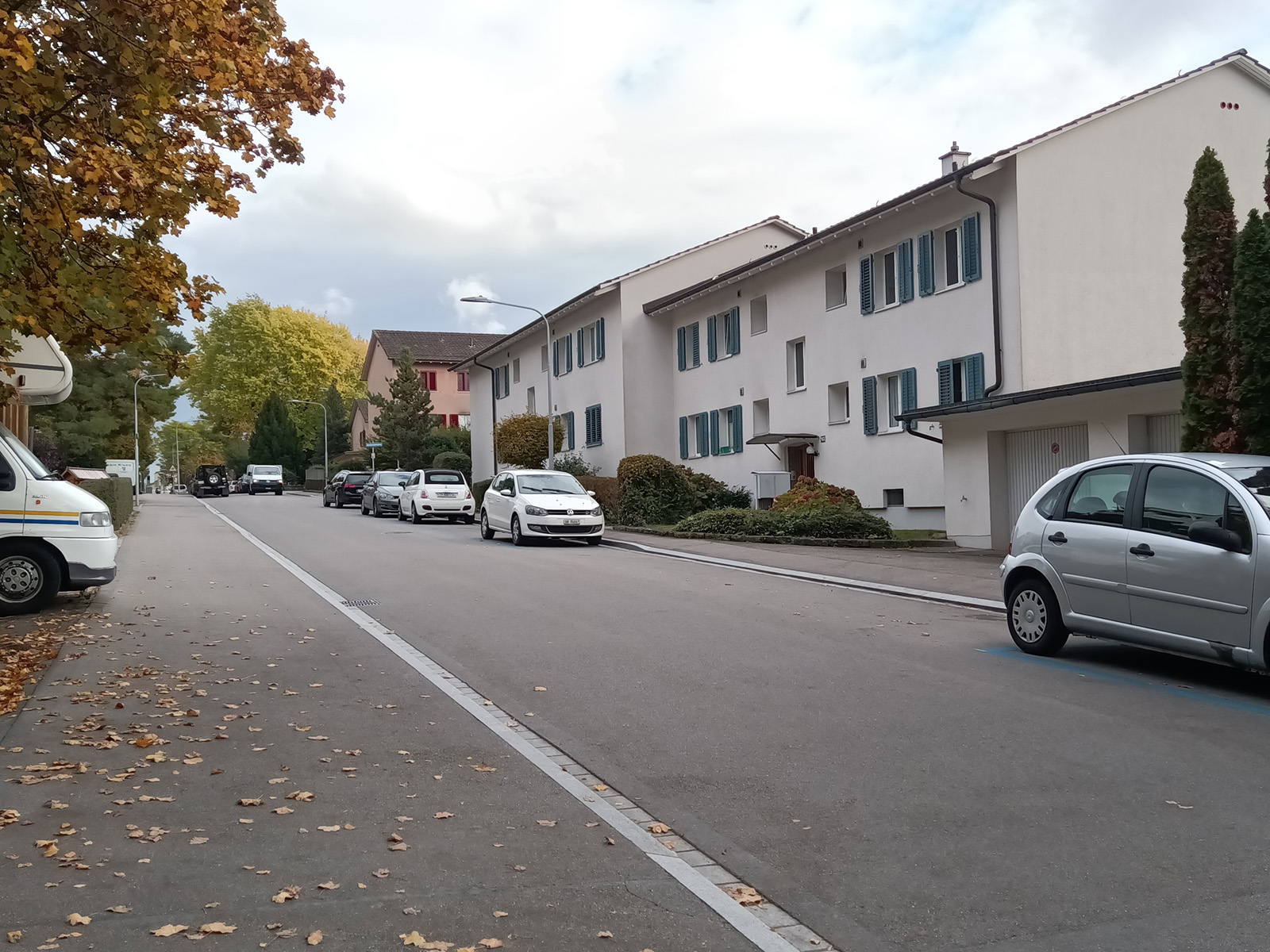}&
    \includegraphics[width=0.24\linewidth]{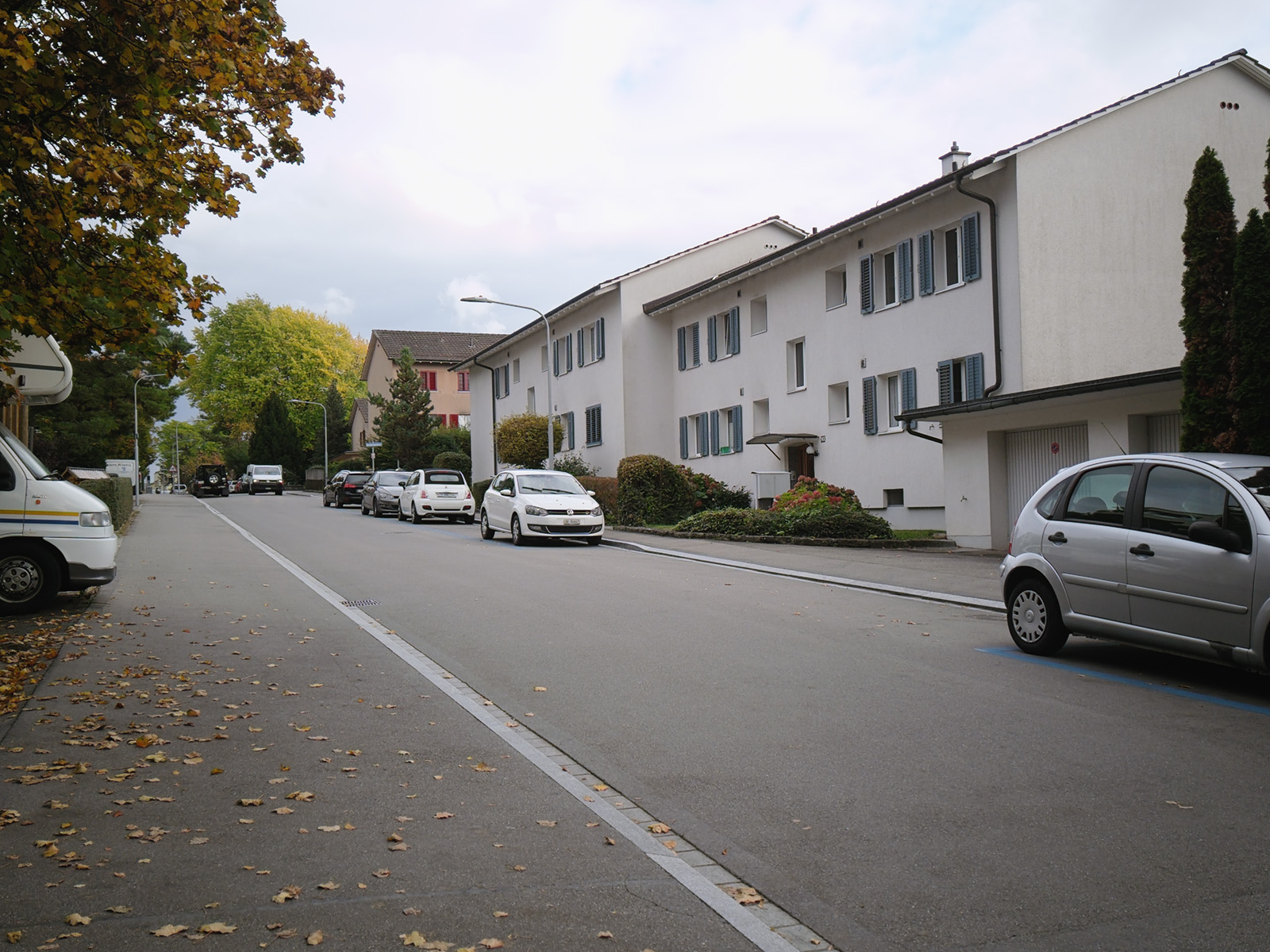}&
    \includegraphics[width=0.24\linewidth]{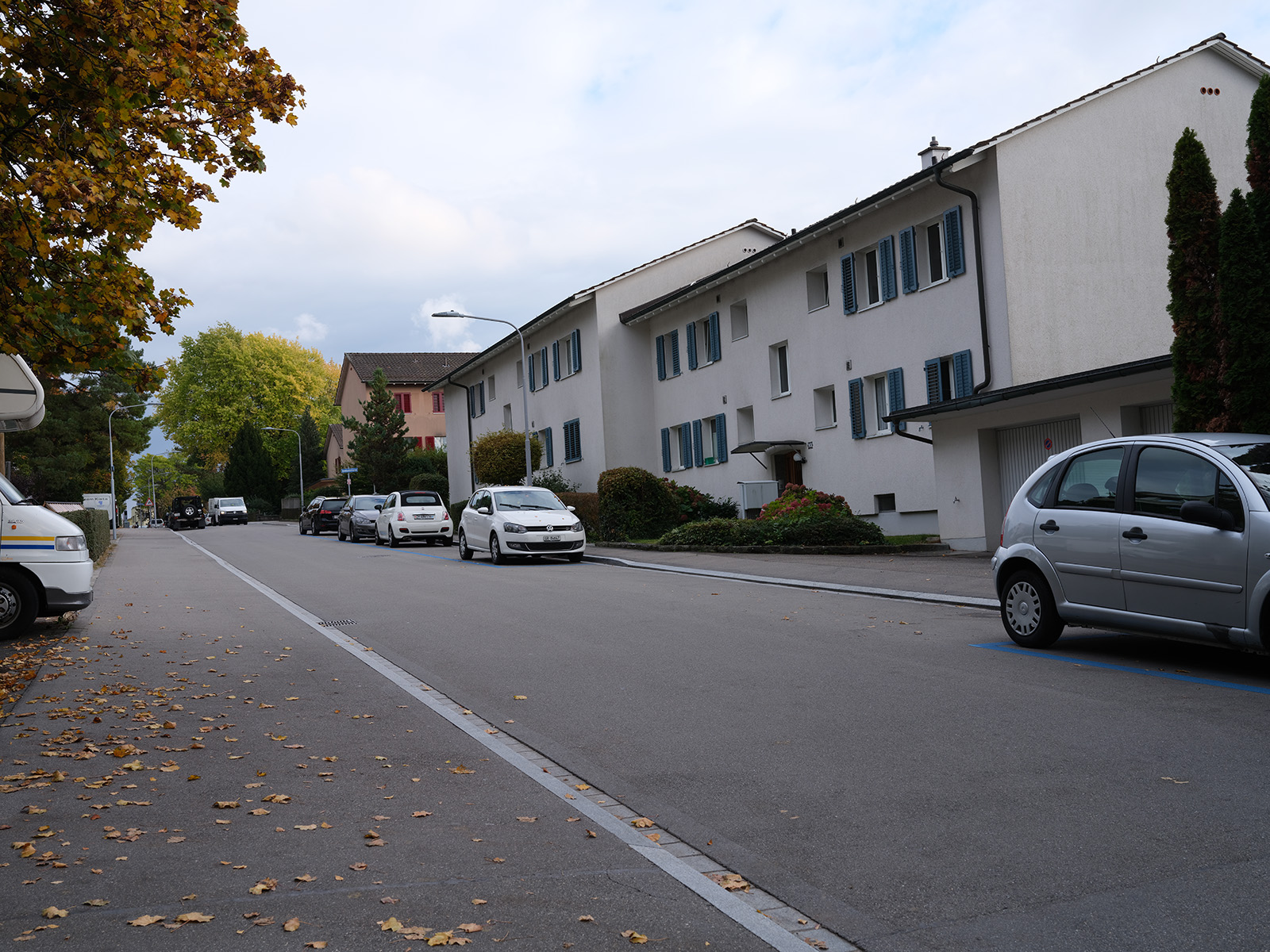} \\
    \includegraphics[width=0.24\linewidth]{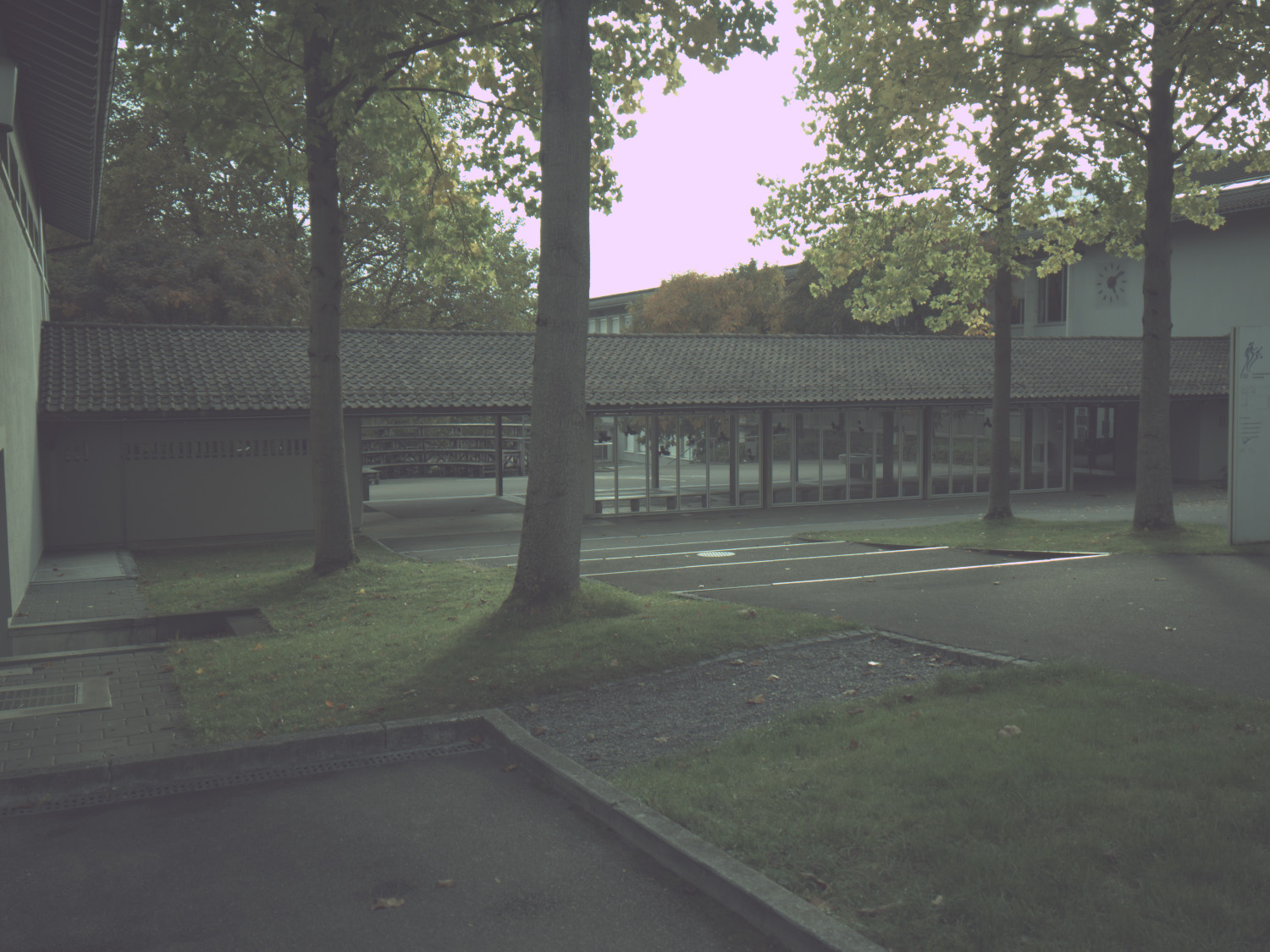}&
    \includegraphics[width=0.24\linewidth]{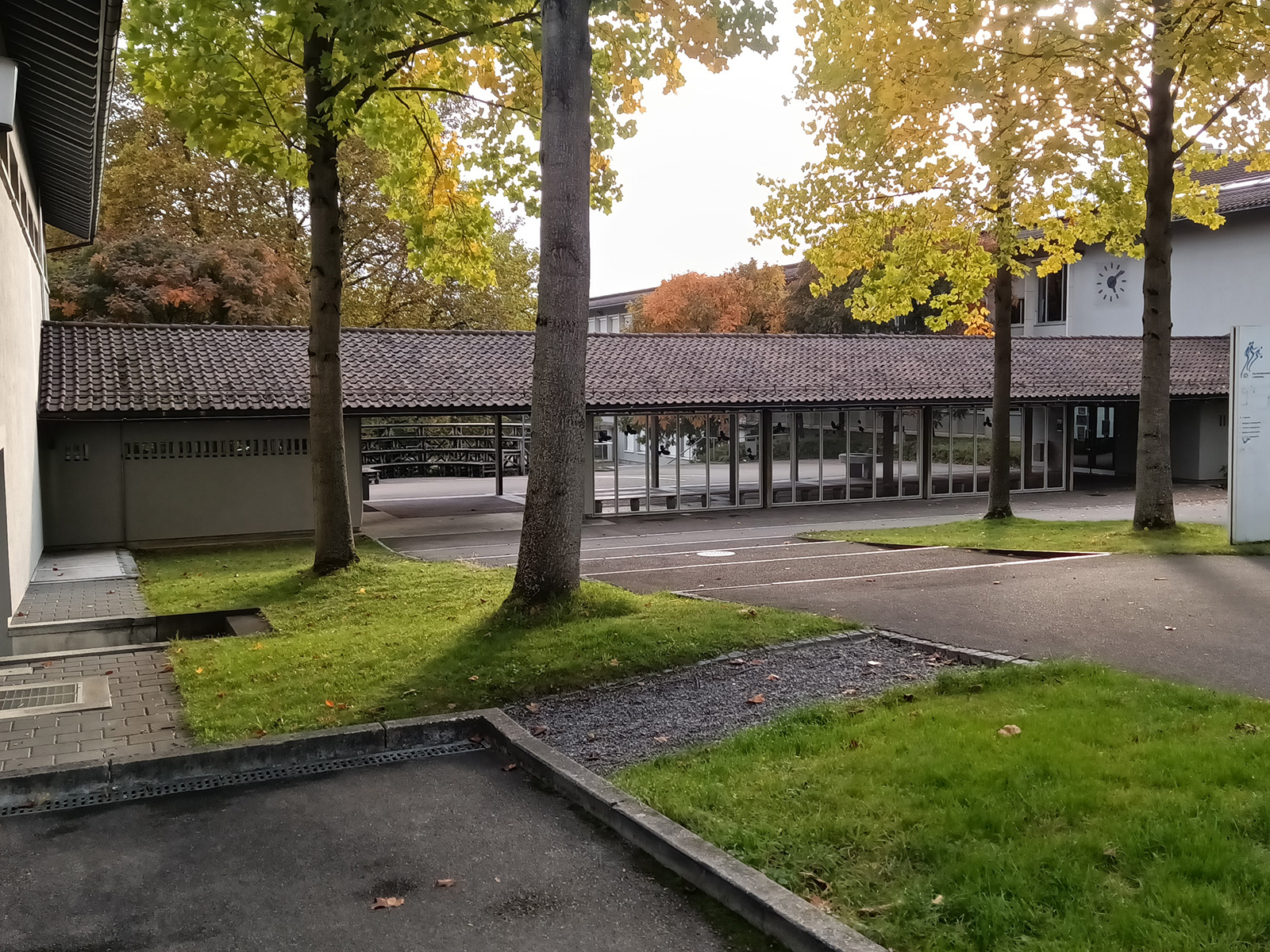}&
    \includegraphics[width=0.24\linewidth]{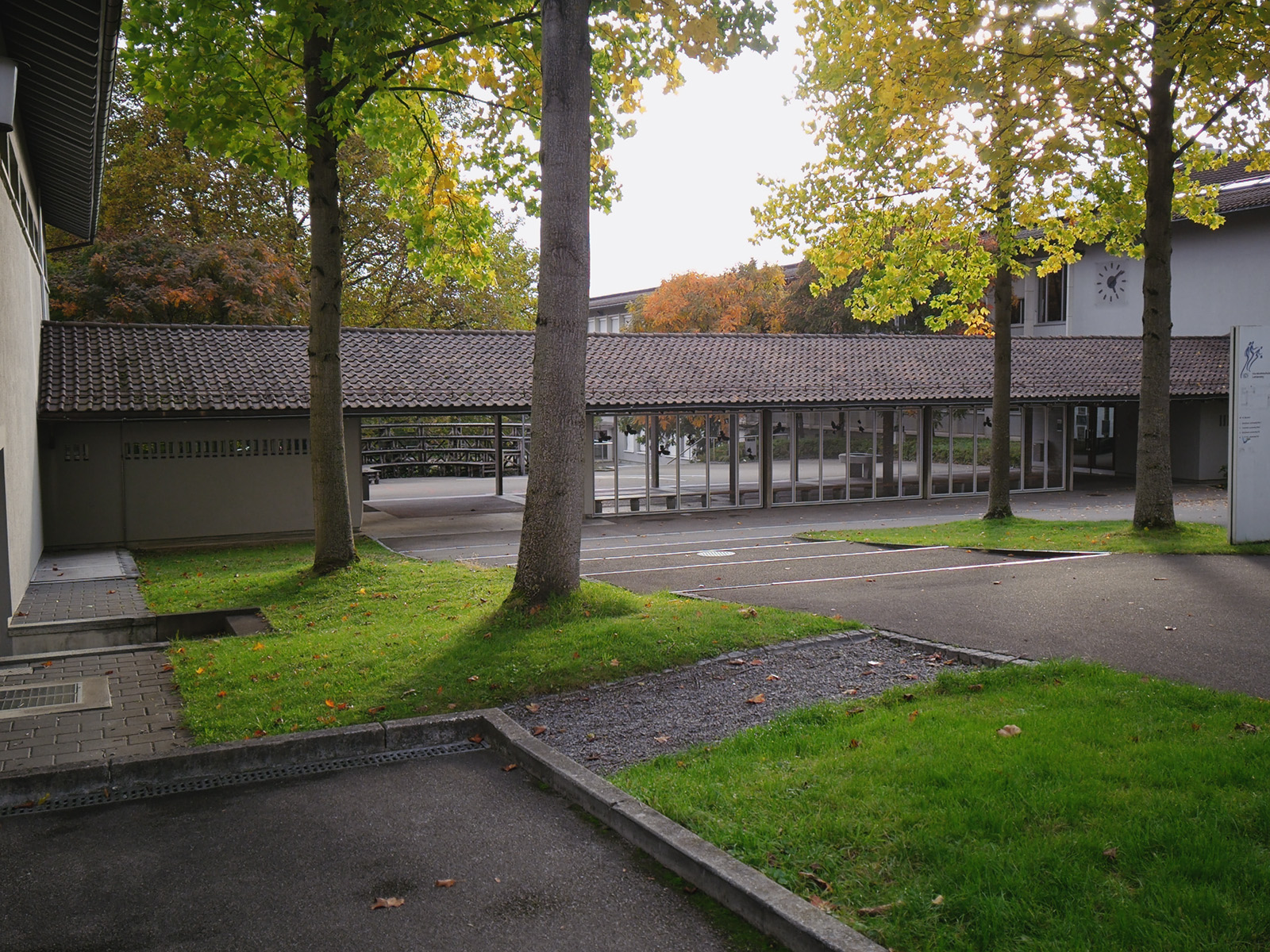}&
    \includegraphics[width=0.24\linewidth]{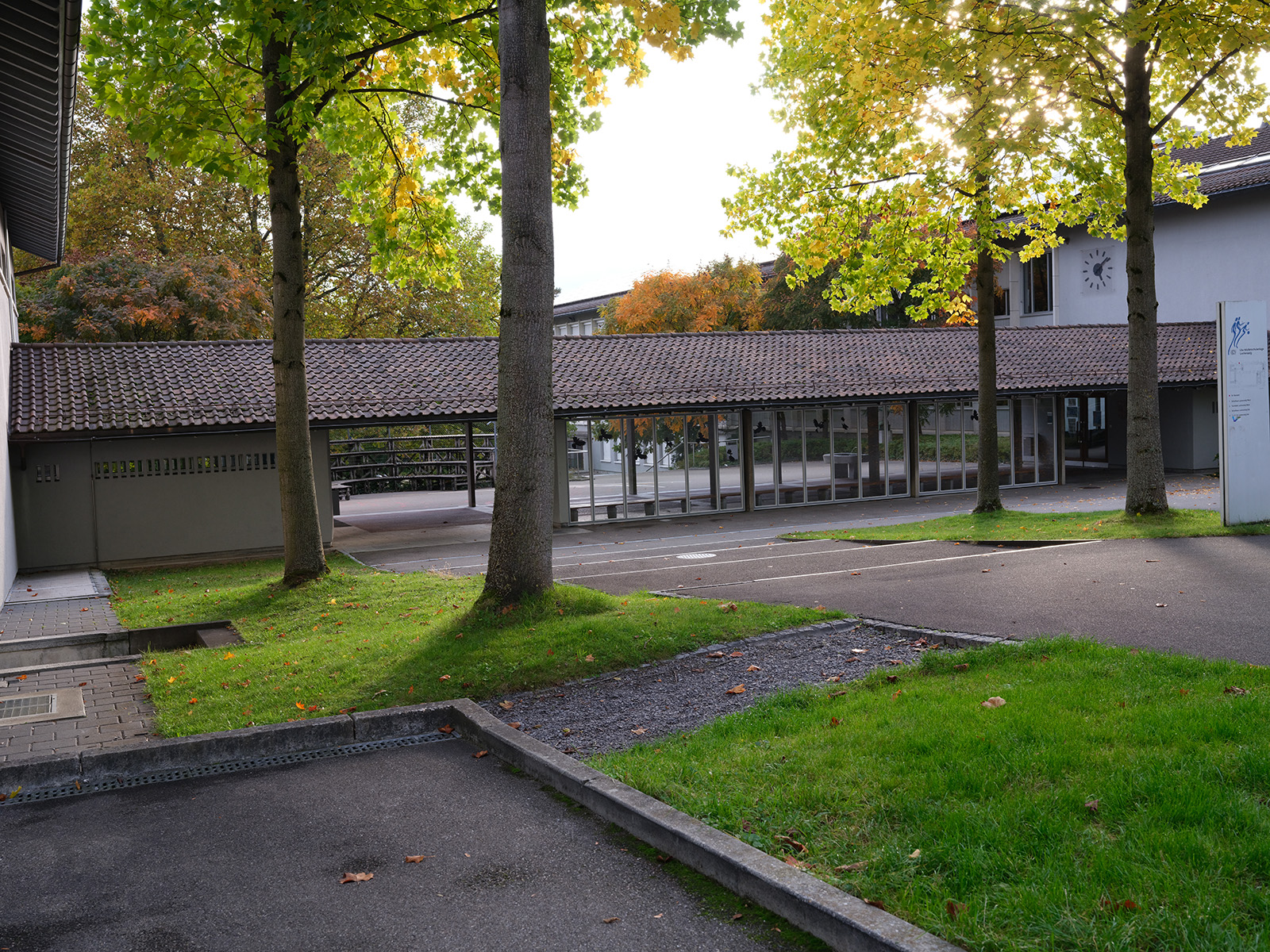} \\
    \includegraphics[width=0.24\linewidth]{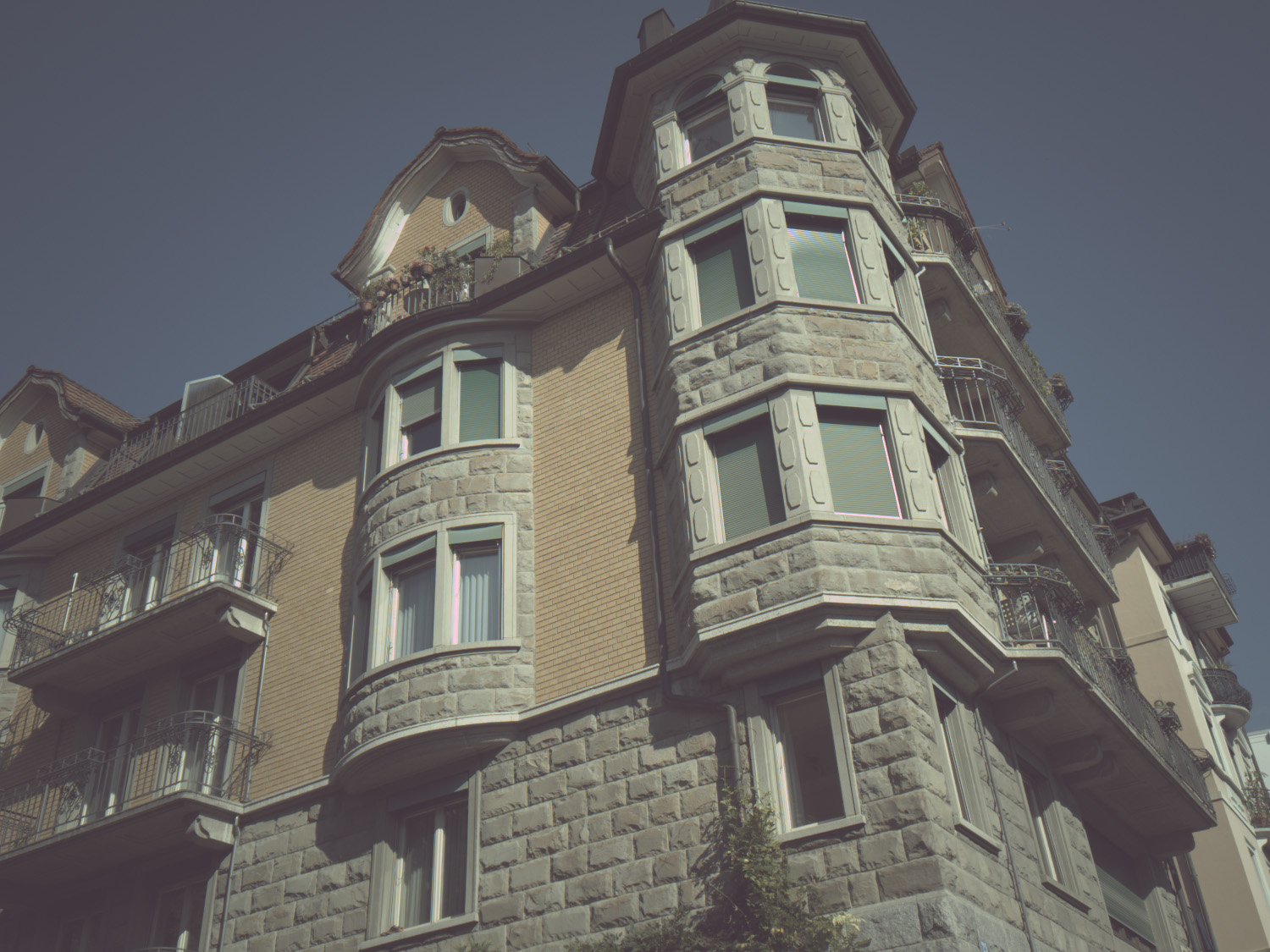}&
    \includegraphics[width=0.24\linewidth]{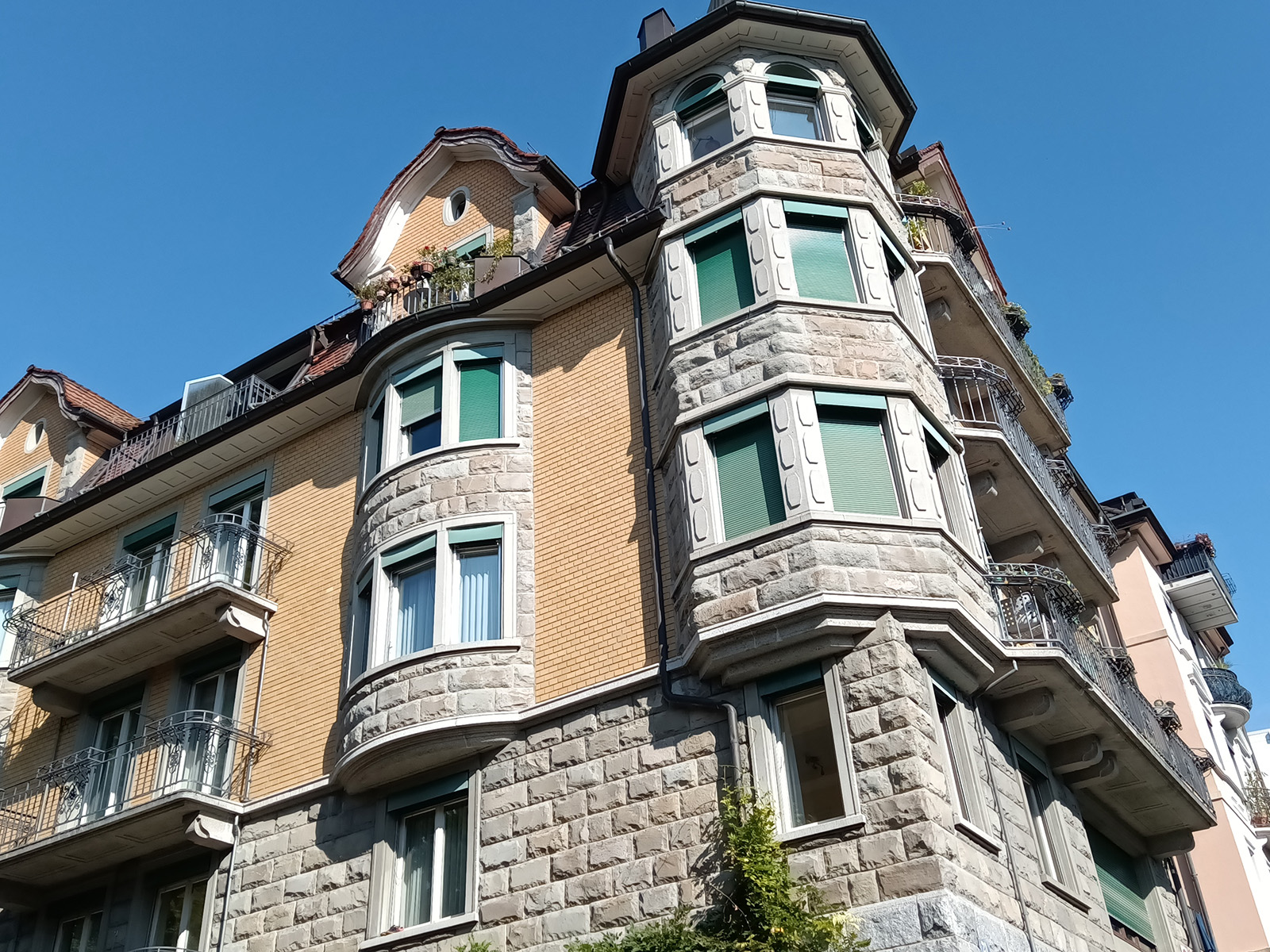}&
    \includegraphics[width=0.24\linewidth]{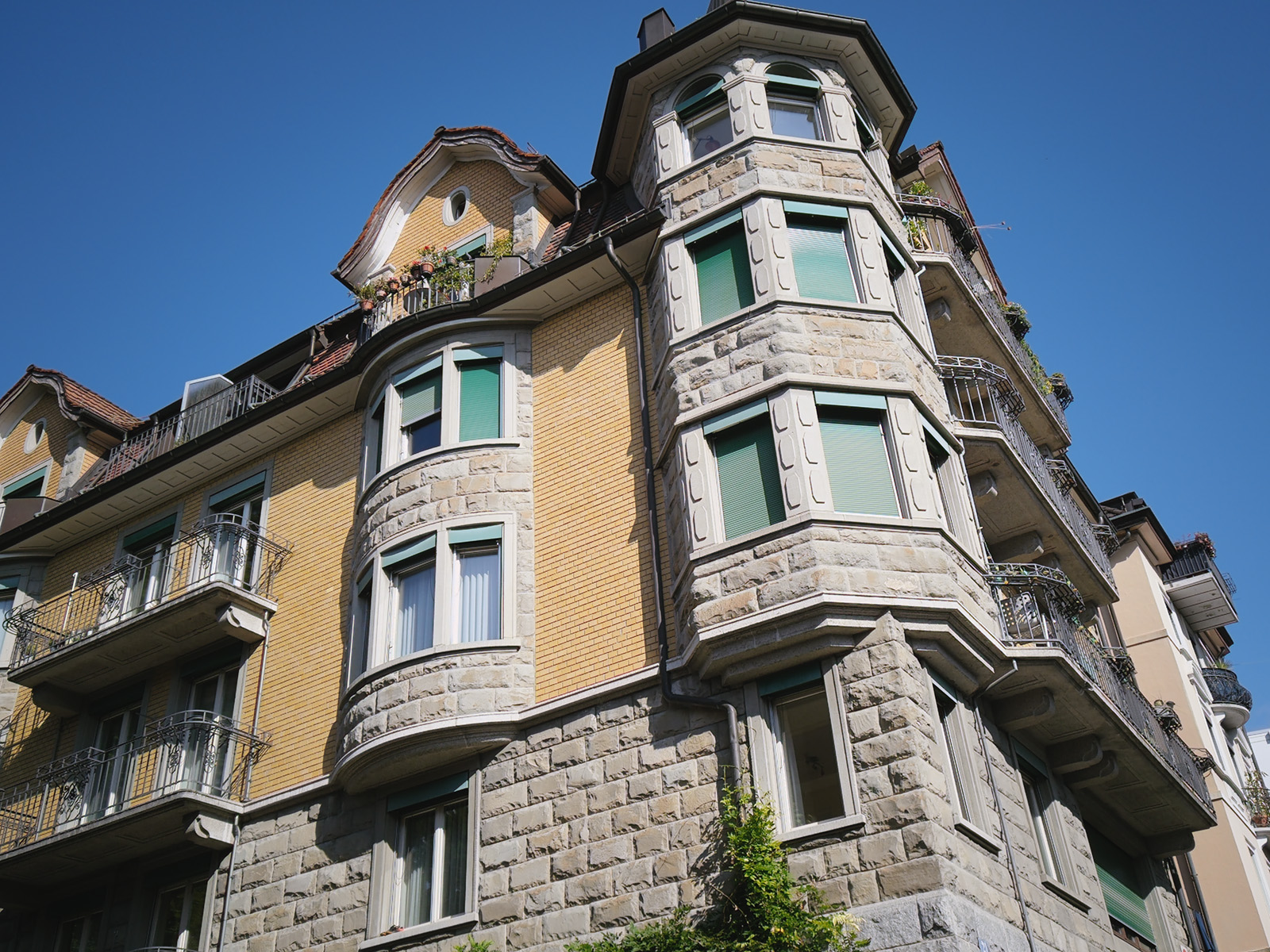}&
    \includegraphics[width=0.24\linewidth]{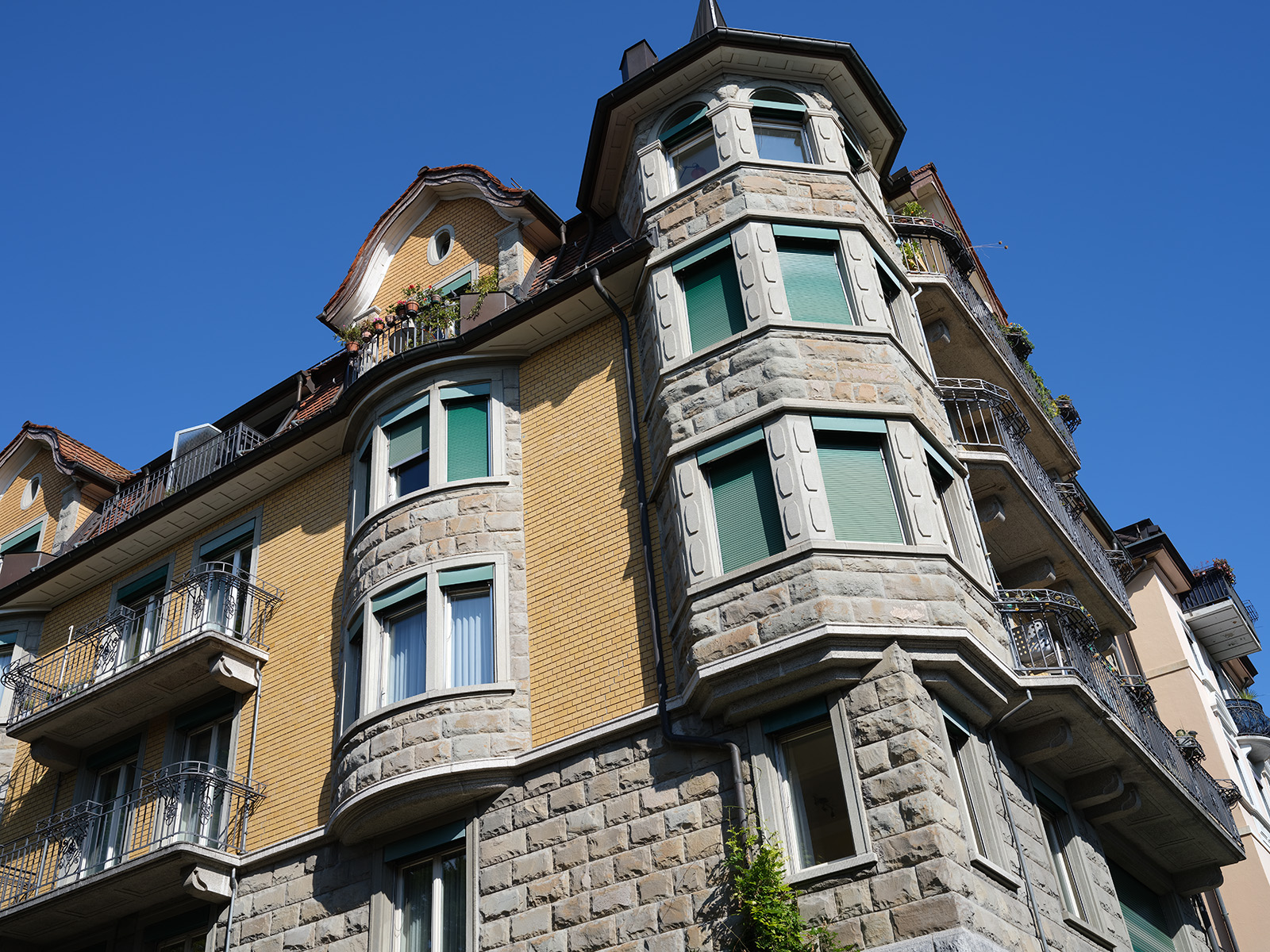} \\
\end{tabular}
}
\vspace{0.0cm}
\caption{Sample visual results obtained with the proposed deep learning method. Best zoomed on screen.}
\label{fig:sample_results}
\end{figure*}

We present a solution that addresses all these limitations while achieving good visual results. The overall architecture of the proposed MicroISP model is illustrated in Fig.~\ref{fig:architecture}. Below we discuss its details and justify our design choices.

\smallskip

\noindent \textBF{Overall image processing workflow.} The model accepts the raw RGBG Bayer data coming directly from the camera sensor. The input is then grouped in 4 feature maps corresponding to each of the four RGBG color channels using the space-to-depth op. Next, this input is processed in parallel in 3 model branches corresponding to the R, G and B color channels and consisting of $N$ residual building blocks (2 blocks are used by default if not specified). After applying the depth-to-space op at the end of each branch, their outputs are concatenated into the reconstructed RGB photo.

\smallskip

\noindent \textBF{Convolutional layers.} When running the model on mobile NPUs or GPUs, the memory is typically allocated per layer / operator. Thus, the maximum RAM consumption is defined by the largest layer that becomes a bottleneck for high-resolution inference. Since the memory consumption for a conv layer is proportional to the number of input/output feature maps $N_{in}/N_{out}$ and their size $H \times W$, our model only uses \textit{4 convolutional filters} in every layer to achieve the minimum possible RAM consumption. This is the smallest possible filter size: as the model takes 4 input RGBG channels, one would need to have at least 4 filters of the same size in the first conv layer to avoid information loss (if pixel shuffle / strided convolution with stride $c$ is applied, then the number of filters should be $4\times c \times c$, leading to the same resulting memory footprint). Each convolutional layer is followed by the \textit{PReLU} activation with shared non-channel dimensions, meaning that only 4 parameters corresponding to each input channel are learned.

\smallskip

\noindent \textBF{Model branches.} The MicroISP model processes the input sensor data in 3 separate branches. First of all, this is done to fit the above mentioned memory constraints while not limiting the performance of the model: if only one branch is used, then 4 convolutional filters would not be enough to perform an accurate image demosaicing and texture reconstruction. Note that in this case one would also need to use 12 filters in the last conv layer since the final depth-to-space op should produce an output image with 3 channels of twice larger resolution (using the transposed convolution instead would lead to serve checkerboard artifacts in the resulting photo, and thus should be avoided).

The other benefit of using separate branches is that this allows the model to learn a different set of features specific for each color space. Indeed, as these branches get the same input data, they can extract and work only with features relevant for reconstructing R, G and B image channels, respectively, while dropping non-relevant information.

Finally, the proposed branch structure is also beneficial from the performance perspective: if the AI accelerator has enough RAM, it can run these branches in parallel as they are independent from each other, and thus the runtime can be decreased by up to 3 times. Alternatively, these branches can be executed sequentially if the resolution of the input photos is too high for parallel data processing.

\smallskip

\noindent \textBF{Attention blocks.} To ensure that the model is able to perform global image processing such as white balancing, gamma and color correction, we added an enhanced channel attention block with the structure of Fig.~\ref{fig:architecture}. While the standard attention units are using global average pooling followed by several conv layers, our initial experiments revealed that the performance of these blocks is not sufficient in our case as they are not taking into account any information about the image content that is removed after the pooling layer. Thus, we propose an enhanced structure: first, a $1\times1$ convolution with stride 3 is applied to reduce the dimensionality of the feature maps. Next, three convolutional blocks with $3\times3$ filters and stride 3 are applied to learn the global content-dependent features and reduce the resolution of the feature maps another 27 times. Finally, the average pooling op is used to get $1\times 1 \times 4$ features that are then passed to 2 additional conv layers generating the normalization coefficients. The proposed architecture is both performant and computationally efficient due to aggressive dimensionality reduction, leading to an execution time of the overall attention block appr. twice smaller than the runtime of one normal $3\times 3$ convolution.

\smallskip

\begin{figure*}[t!]
\centering
\setlength{\tabcolsep}{1pt}
\resizebox{\linewidth}{!}
{
\begin{tabular}{ccccccc}
    \includegraphics[width=0.2\linewidth]{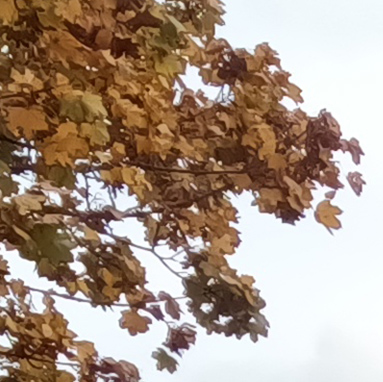}&
    \includegraphics[width=0.2\linewidth]{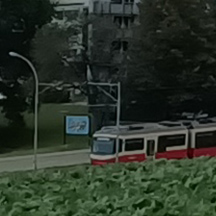}&
    \includegraphics[width=0.2\linewidth]{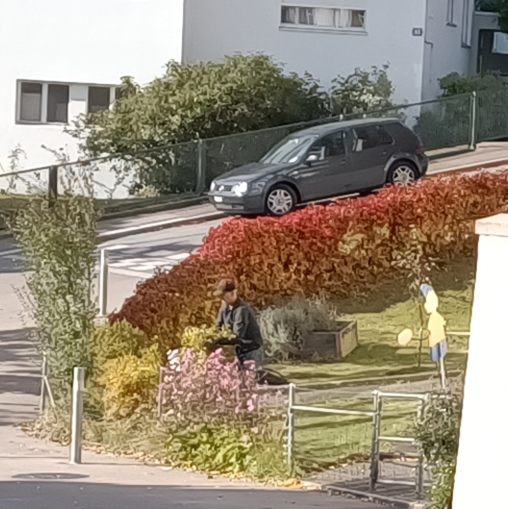}&
    \includegraphics[width=0.2\linewidth]{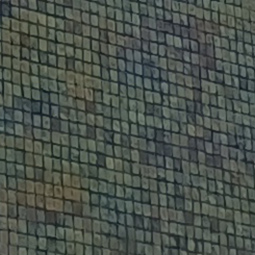}&
    \includegraphics[width=0.2\linewidth]{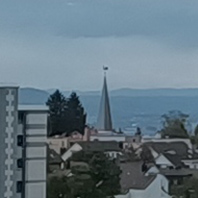}&
    \includegraphics[width=0.2\linewidth]{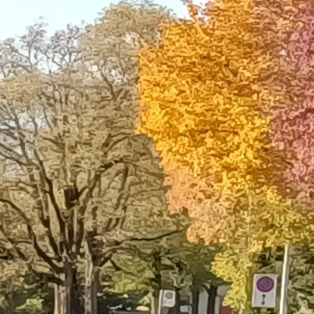}&
    \includegraphics[width=0.2\linewidth]{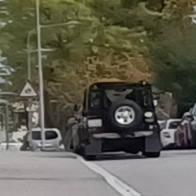}\\
    \includegraphics[width=0.2\linewidth]{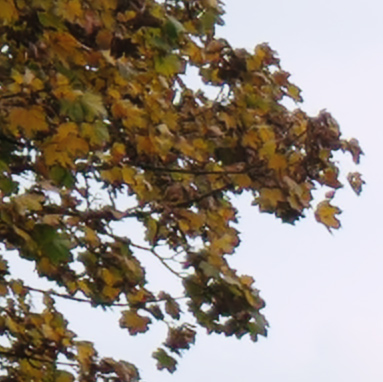}&
    \includegraphics[width=0.2\linewidth]{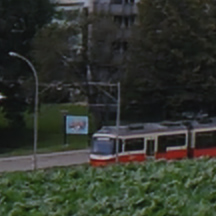}&
    \includegraphics[width=0.2\linewidth]{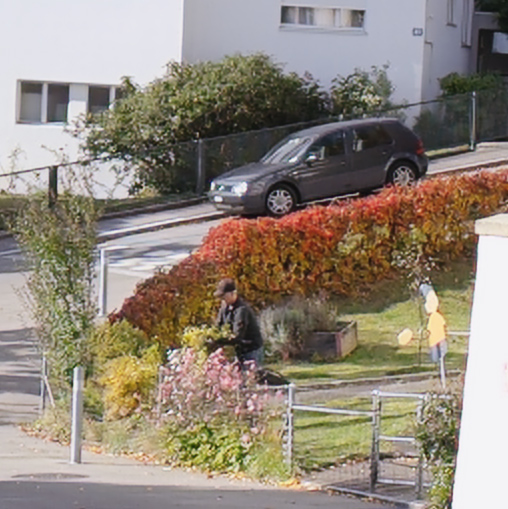}&
    \includegraphics[width=0.2\linewidth]{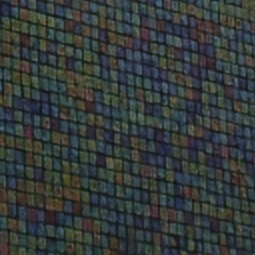}&
    \includegraphics[width=0.2\linewidth]{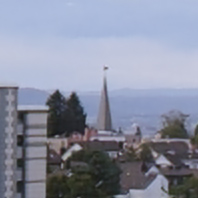}&
    \includegraphics[width=0.2\linewidth]{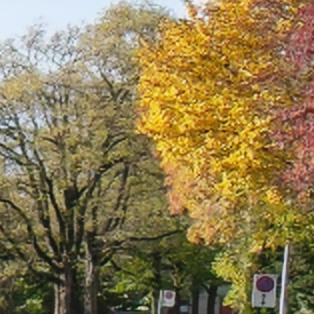}&
    \includegraphics[width=0.2\linewidth]{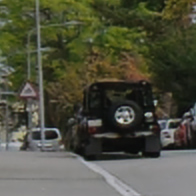}\\
\end{tabular}
}
\vspace{-0.2cm}
\caption{Sample crops from the MediaTek ISP photos (top) and the corresponding images processed with the MicroISP model (bottom).}
\vspace{-0.2cm}
\label{fig:crops}
\end{figure*}

\noindent \textBF{Model operators.} The proposed MicroISP model contains only layers supported by the Neural Networks API 1.2~\cite{NNAPI12Specs}, and thus can run on any NNAPI-compliant AI accelerator (such as NPU, APU, DSP or GPU) available on mobile devices with Android 10 and above. One can additionally relax the above requirements to NNAPI 1.0 and Android 8.1, resp., if replacing \textit{PReLU} with \textit{Leaky ReLU}. The effect of this change is discussed in the next section.

\smallskip

\noindent \textBF{Model size and memory consumption.} The size of the MicroISP network is only 158 KB when exported for inference using the TFLite FP32 format. The model consumes around 90, 475 and 975MB of RAM when processing FullHD, 12MP and 32MP photos on mobile GPUs, respectively.

\smallskip

\noindent \textBF{Training details.} The model was trained in three stages. First, only the MSE loss was used for 200 epochs to get the initial reconstruction results. Next, the model was tuned for another 200 epochs with a combination of the VGG-based~\cite{johnson2016perceptual} perceptual, SSIM and MSE losses to improve the texture quality, enhance the details and image sharpness:
\begin{equation*}
\label{eq:loss}
\mathcal{L}_{\text{Stage 2}} = \mathcal{L}_{\text{VGG}} + 0.5 \cdot \mathcal{L}_{\text{SSIM}} + 0.25 \cdot \mathcal{L}_{\text{MSE}},
\end{equation*}
where the value of each loss is normalized to 1. Finally, the network was fine-tuned for another 100 epochs with a combination of the SSIM and MSE loss functions taken in the ratio of 2:1. This was done to perform the final tone mapping adjustments and to improve edge rendering.

\smallskip

\noindent \textBF{Implementation details.} The model was implemented in TensorFlow and trained on a single \textit{Nvidia Titan X} GPU with batch size 50. The network parameters were optimized for 500 epochs using the ADAM~\cite{kingma2014adam} algorithm with a learning rate of $2e^{-5}$. Random flips and rotations were applied to augment the training data and prevent overfitting.

\section{Experiments}
\label{sec:experiments}

In this section, we evaluate the proposed MicroISP architecture on the real Fujifilm UltraISP dataset and mobile devices to answer the following questions:
\begin{itemize}
\itemsep -0.12em
\item Is the model able to perform an accurate reconstruction of the RGB images;
\item How good are the results compared to the standard hand-crafted ISP pipelines used in modern phones;
\item How well this solution performs compared to the commonly used deep learning models tuned for this task;
\item What is the largest image resolution that can be processed by the MicroISP model on mobile devices;
\item What is the runtime of this model when performing the inference on mobile GPUs and AI accelerators,
\item What are the limitations of the proposed solution.
\end{itemize}
To answer these questions, we performed a wide range of experiments which results are described in detail in the following sections.

\subsection{Qualitative evaluation}

As the perceptual quality of the reconstructed photos is our primary target, we started the experiments with a brief analysis of the visual results obtained with the MicroISP model. Figure~\ref{fig:sample_results} shows sample RGB images reconstructed with the proposed solution together with the original RAW photos, images obtained with MediaTek's built-in ISP system, and the target photos from the Fujifilm camera. The first observation demonstrates that the results produced by the model are valid: it was able to perform an accurate color reconstruction with decent tone mapping, the quality of texture and the overall image sharpness are quite good, white balancing is performed correctly. The same can be also applied to the dynamic range that is close to the one on the target Fujifilm images. No notable issues or artifacts are observed at both local and global levels, complex overexposed image areas are also handled correctly.

Surprisingly, the images rendered with the neural network turned out to look more natural compared to the ones obtained with the built-in ISP pipeline. A more detailed analysis of image crops revealed that this is mainly caused by a strong watercolor effect present on the majority of photos processed with the ISP system (Fig.~\ref{fig:crops}). The reason for this is that most modern smartphones apply numerous filters used for image sharpening and low-level texture enhancement, though together with aggressive noise suppression algorithms they are often leading to a mess of pixels in complex image areas such as grass or leaves, and a notable loss of colors. Overall, the photos obtained with the ISP system and with the proposed solution are following the two considerably different approaches: the first ones have boosted colors, a significantly increased brightness and lots of sharpening used to make them visually more appealing, while the images processed with the MicroISP model are looking naturally as one would expect them to be. The real resolution of the ISP images is slightly higher for some scenes, though the difference is overall quite negligible. It should be also noted that, as expected, the target Fujifilm photos significantly outperform the results of the ISP pipeline and the MicroISP model in all aspects, especially in terms of resolution.

\begin{figure*}[t!]
\centering
\setlength{\tabcolsep}{1pt}
\resizebox{\linewidth}{!}
{
\begin{tabular}{cccc}
    \includegraphics[width=0.2\linewidth]{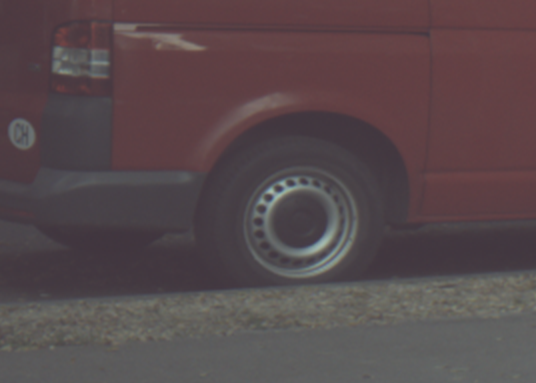}&
    \includegraphics[width=0.2\linewidth]{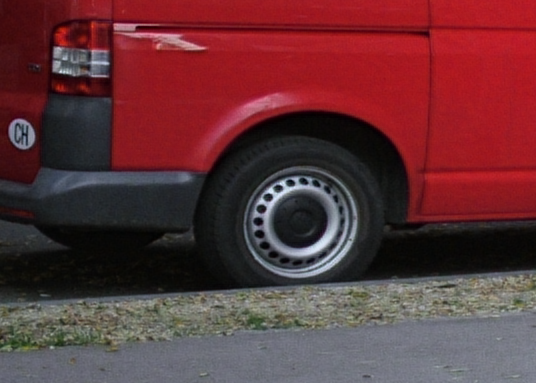}&
    \includegraphics[width=0.2\linewidth]{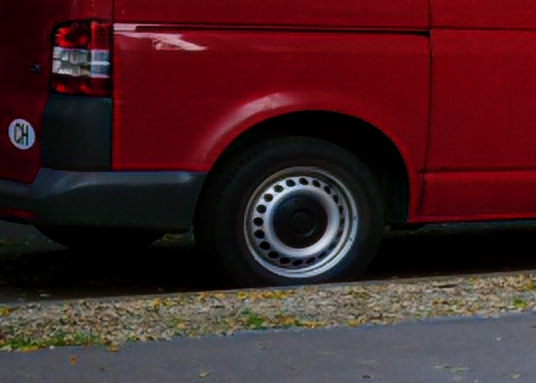}&
    \includegraphics[width=0.2\linewidth]{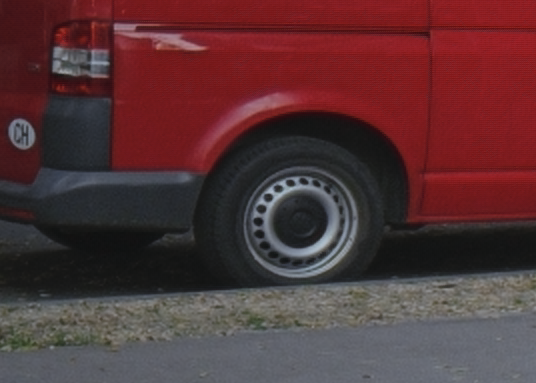}\\
    \includegraphics[width=0.2\linewidth]{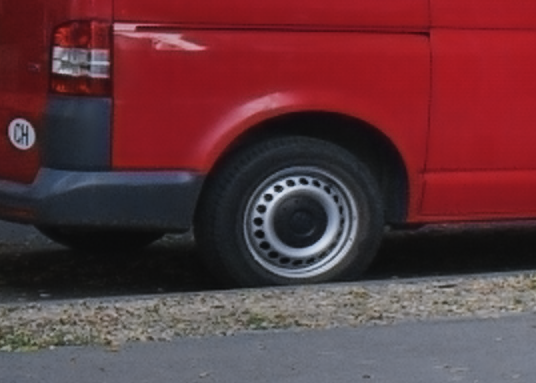}&
    \includegraphics[width=0.2\linewidth]{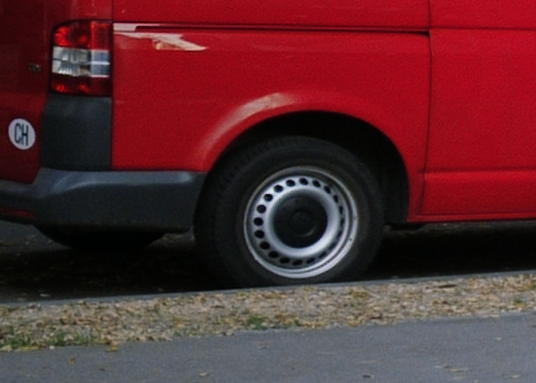}&
    \includegraphics[width=0.2\linewidth]{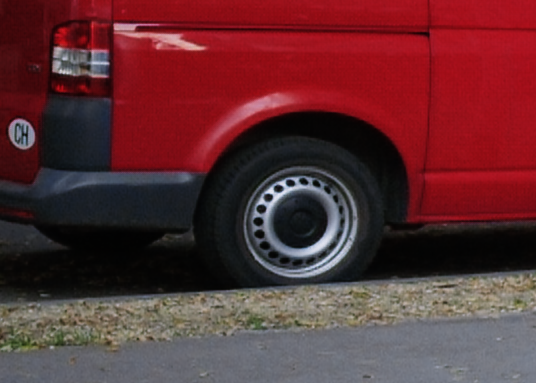}&
    \includegraphics[width=0.2\linewidth]{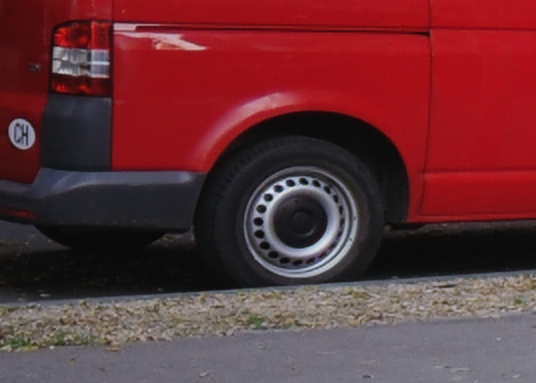}\\
\end{tabular}
}
\vspace{-0.2cm}
\caption{From left to right, top to bottom: the original visualized cropped RAW image, and the same image after applying, respectively: SmallNet~\cite{ignatov2021learned}, FPIE~\cite{de2018fast}, FSRCNN~\cite{dong2016accelerating}, Compressed UNet~\cite{ignatov2021learned}, ENERZAi~\cite{ignatov2021learned}, CSANet~\cite{hsyu2021CSANet} and our MicroISP model.}
\vspace{-0.2cm}
\label{fig:models}
\end{figure*}

\subsection{Quantitative evaluation}

As the proposed solution was designed targeting good visual results and fast on-device high resolution image processing, in the next two sections we compare its numerical and runtime results against the previously introduced deep learning based approaches allowing to perform RGB image reconstruction directly on smartphones. The following models are used in the next experiments:
\begin{itemize}
\itemsep -0.12em
\item FSRCNN~\cite{dong2016accelerating}: a popular computationally efficient model used for various image enhancement problems.
\item FPIE~\cite{de2018fast}: an enhanced DPED~\cite{ignatov2017dslr}-based neural network optimized for fast on-device image processing.
\item CSANet~\cite{hsyu2021CSANet}: an NPU-friendly architecture developed for the learned smartphone ISP problem.
\item Compressed U-Net~\cite{ignatov2021learned}: a U-Net~\cite{ronneberger2015u} based model with hardware-specific adaptations for edge inference.
\item ENERZAi~\cite{ignatov2021learned}: a model designed for efficient image ISP, derived from the ESRGAN~\cite{wang2018esrgan} architecture.
\item SmallNet~\cite{ignatov2021learned}: a fast FSRCNN~\cite{dong2016accelerating} based architecture optimized for the learned smartphone ISP task.
\end{itemize}
All models were trained on the Fujifilm UltraISP dataset, their PSNR and SSIM scores on the test image subset are reported in Table~\ref{tab:numerical_results}, sample visual results for all methods are demonstrated in Fig.~\ref{fig:models}. The proposed MicroISP network was able to substantially outperform the other solutions in almost all aspects. In particular, it offers a 0.5dB PSNR improvement compared to the baseline FSRCNN model, and outperforms by 0.14 dB the CSANet model demonstrating the second best PSNR results on the considered dataset. When analyzing the visual results, one can note that, unlike the majority of other solutions, it produces images with smooth and clear texture and without checkerboard artifacts that are especially severe on the photos processed by the FPIE and CSANet networks. Additionally, this is the only model that also learned to perform image denoising~--- an aspect that is critical when processing mobile photos, where noise is often present even on images captured in good lighting conditions.

\subsection{Runtime evaluation}

As the proposed solution is designed for on-device image processing, we perform its performance evaluation directly on mobile phones to get the real runtime values and take into account all limitations related to edge inference. For this, we used the publicly available AI Benchmark application~\cite{ignatov2018ai,ignatov2019ai} that allows to load any custom TensorFlow Lite model and run it on any Android device with various acceleration options including GPU, NPU / DSP and CPU inference. Same as in~\cite{ignatov2021learned}, we used the MediaTek Dimensity 1000+ mobile SoC for performing runtime evaluation, and accelerated the models on its Mali-G77 GPU as this option delivered the best latency for all architectures. The results of all models on FullHD, 12MP, 18MP, 26MP and 32MP photos are reported in Table~\ref{tab:runtime_comparison}.

\begin{table}[t!]
\centering
\resizebox{0.8\linewidth}{!}
{
\begin{tabular}{l|cc}
\hline
Method \, & \, PSNR \, & \, SSIM \, \\
\hline
\hline
SmallNet~\cite{ignatov2021learned} & 23.20 & 0.847 \\
FPIE~\cite{de2018fast} & 23.23 & 0.848 \\
FSRCNN~\cite{dong2016accelerating} & 23.27 & 0.830 \\
Compressed U-Net~\cite{ignatov2021learned} & 23.30 & 0.840 \\
ENERZAi~\cite{ignatov2021learned} & 23.41 & \textBF{0.853} \\
CSANet~\cite{hsyu2021CSANet} & 23.73 & 0.849 \\
\hline
MicroISP & \textBF{23.87} & \textBF{0.853} \\
\end{tabular}
}
\vspace{-0.5mm}
\caption{Average PSNR / SSIM results on test images.}
\label{tab:numerical_results}
\vspace{-1.5mm}
\end{table}

\begin{table}[t!]
\centering
\resizebox{1.0\linewidth}{!}
{
\begin{tabular}{l|ccccc}
\hline
Method \, & \multicolumn{5}{|c}{Runtime on the Dimensity 1000+ GPU} \\
& \, FullHD, \, & \, 12MP, \, & \, 18MP, \, & \, 26MP, \, & \, 32MP, \\
& \, ms \, & ms & ms & ms & ms \\
\hline
\hline
SmallNet~\cite{ignatov2021learned}          & 18.4  & 100   & 148 & \small{OOM} & \small{OOM} \\
FPIE~\cite{de2018fast}                      & 208   & 1138  & \small{OOM} & \small{OOM} & \small{OOM} \\
FSRCNN~\cite{dong2016accelerating}          & 40.8  & 232   & 335 & \small{OOM} & \small{OOM} \\
Compressed U-Net~\cite{ignatov2021learned}   & 29.1  & 140   & 208 & \small{OOM} & \small{OOM} \\
ENERZAi~\cite{ignatov2021learned}           & 31.2  & 123   & 184 & \small{OOM} & \small{OOM} \\
CSANet~\cite{hsyu2021CSANet}                & 44.2  & 241   & 358 & \small{OOM} & \small{OOM} \\
\hline
DPED~\cite{ignatov2017dslr}                 & 658   & 4027  & \small{OOM} & \small{OOM} & \small{OOM} \\
PyNET~\cite{ignatov2020replacing}           & 12932 & \small{OOM}   & \small{OOM} & \small{OOM} & \small{OOM} \\
\hline
MicroISP                                    & 42.3  & 238   & 354 & 522 & 636 \\
\end{tabular}
}
\vspace{-0.2mm}
\caption{The runtime of different deep learning-based models on the MediaTek Dimensity 1000+ GPU obtained using the publicly available AI Benchmark application~\cite{ignatov2018ai,ignatov2019ai}. The results of the DPED and PyNET models are provided for the reference. OOM stands for the ``out-of-memory'' exception thrown by the interpreter when trying to perform the inference.}
\label{tab:runtime_comparison}
\vspace{-4mm}
\end{table}

\begin{table*}[t!]
\centering
\resizebox{\linewidth}{!}
{
\begin{tabular}{l|cc|cc|cc|c}
\hline
Mobile SoC & \,  Dimensity 9000 \,  & \,  Dimensity 820 \,  & \,  Exynos 2100 \,  & \, Exynos 990 \, & \, Kirin 9000 \, & \, Snapdragon 888 \, & \, Google Tensor \\
GPU & \, \small Mali-G710 MC10, ms \, & \, \small Mali-G57 MC5, ms \, & \, \small Mali-G78 MP14, ms \, & \, \small Mali-G77 MP11, ms \, & \, \small Mali-G78 MP24, ms \, & \, \, \small Adreno 660, ms \, & \, Mali-G78 MP20, ms \\
\hline
Full HD & 33.4 & 72.7 & 31.2 & 46.2 & 39.4 & 30.4 & 36.8 \\
12MP    & 170 & 416 & 174  & 248 & 189 & 153 & 188 \\
32MP    & 624 & 1059 & 489 & 690 & 507 & 465 & 480 \\
\end{tabular}
}
\vspace{-0.2cm}
\caption{The speed of the proposed MicroISP architecture on several popular mobile GPUs for different photo resolutions. The runtime was measured with the AI Benchmark app using the TFLite GPU delegate~\cite{lee2019device}.}
\label{tab:runtime_on_socs}
\vspace{-0.2cm}
\end{table*}

As expected, the MicroISP network was able to achieve the lowest memory consumption, and thus was the only architecture capable of processing 26MP and 32MP images, while in all other cases the TensorFlow interpreter failed to perform the inference with the \textit{out-of-memory} exception. The model was able to achieve a runtime of 42.3, 238 and 354 ms on FullHD, 12MP and 18MP photos, respectively, which is comparable to the latency of the FSRCNN and CSANet models, while the proposed solution provides better numerical and visual results. Though the SmallNet, Compressed U-Net and ENERZAi models are significantly faster in this experiment, in section~\ref{sec:complexity} we will demonstrate that one can easily achieve a similar runtime and fidelity results by adjusting the number of MicroISP's building blocks. In Table~\ref{tab:runtime_comparison}, we additionally provide the latency of the DPED~\cite{ignatov2017dslr} and PyNET~\cite{ignatov2020replacing} models: despite achieving very good numerical results on the Fujifilm UltraISP dataset (24.2 and 25dB PSNR, respectively), it is almost infeasible to run them on the latest mobile devices due to their huge runtime even on FullHD and 12MP images (more than 15x and 300x higher compared to the MicroISP model).

Finally, we checked the runtime of the proposed solution on all popular mobile chipsets, and report the obtained results in Table~\ref{tab:runtime_on_socs}. When processing FullHD images, the MicroISP model was able to achieve a speed of 30 frames per second on the Dimensity 9000, Exynos 2100 and Snapdragon 888 platforms, which demonstrates that it can be potentially used for real-time FullHD RAW video processing. The results also show that it can perform 32MP photo rendering under 0.5 second on almost all flagship SoCs, though even for mid-range platforms like the MediaTek Dimensity 820 its latency remain reasonably small.

\subsection{Inference on Mobile NPUs}

\begin{table}[b!]
\centering
\resizebox{1.0\linewidth}{!}
{
\begin{tabular}{l|ccccc}
\hline
 \, & \multicolumn{5}{c}{MediaTek Dimensity 9000 mobile chipset} \\
& \, CPU \, & GPU (FP16) & APU (FP16) & APU (INT16) & APU (INT8) \\
\hline
\hline
Runtime, ms & 242 & 33 & 63 & 43 & \textBF{29} \\
Power, fps/watt\, & 0.43 & 5.27 & 7.20 & 7.78 & \textBF{16.61} \\
\end{tabular}
}
\vspace{0.8mm}
\caption{The runtime and power consumption of the MicroISP model on the MediaTek Dimensity 9000 mobile platform.}
\label{tab:apu_performance}
\vspace{-1.2mm}
\end{table}

While the proposed MicroISP model can achieve good runtime results for images of different resolutions on mobile GPUs, one might be interested in running it on dedicated AI accelerators to further improve its latency or to efficiently decrease the power consumption. For this, in this section we evaluate its performance on MediaTek's latest mobile platform, \emph{Dimensity 9000}, which features a powerful AI Processing Unit (APU) designed specifically for complex computer vision and image processing tasks. Table~\ref{tab:apu_performance} shows the runtime and power consumption results obtained on this chipset for FullHD images when running the MicroISP model on CPU, GPU and APU. The considered AI accelerator was able to execute the entire floating-point model without any partitioning, demonstrating a 35\% and 1500\% increase in power efficiency compared to GPU and CPU inference, respectively. These numbers are further increased when quantizing and converting the model to INT16 and INT8 formats. In the latter case, the speed improves from 30 to 34 FPS, and the energy consumption decreases by 3 times compared to GPU execution.

\subsection{Ablation study}

\begin{figure*}[t!]
\centering
\setlength{\tabcolsep}{1pt}
\resizebox{\linewidth}{!}
{
\begin{tabular}{cccc}
    \small{0.25} & \small{0.5} & \small{1.0} & \small{1.5} \\
    \includegraphics[width=0.24\linewidth]{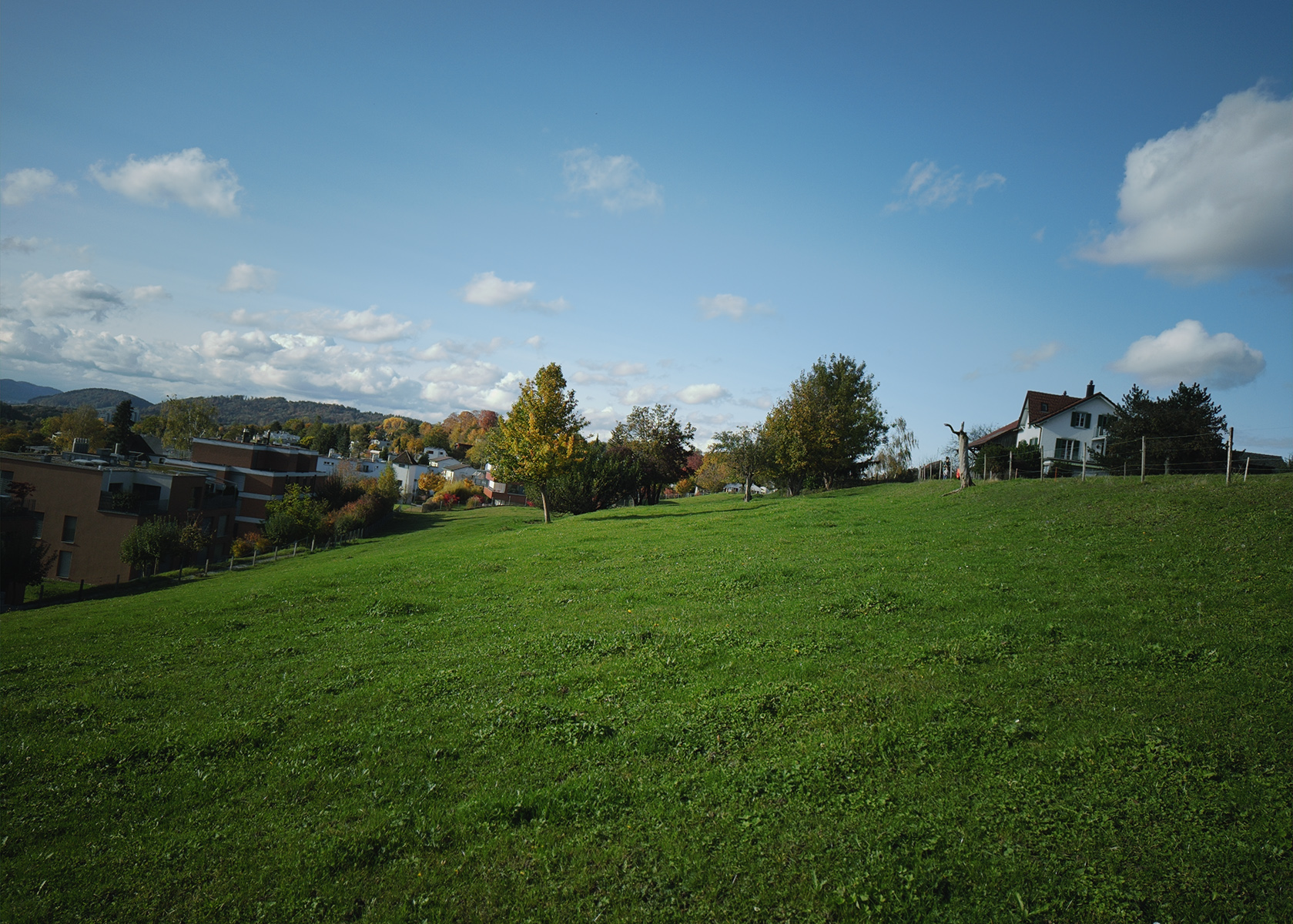}&
    \includegraphics[width=0.24\linewidth]{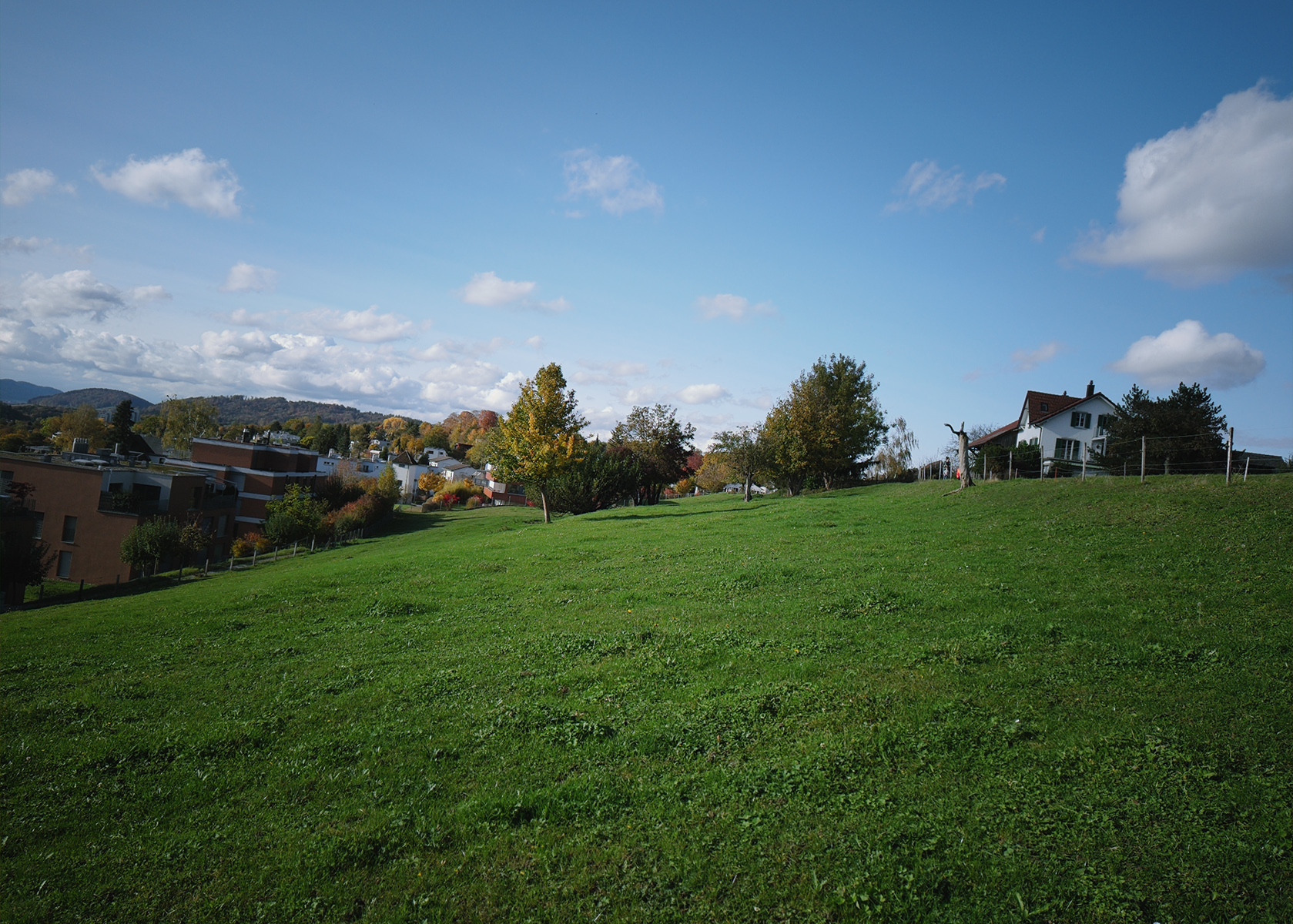}&
    \includegraphics[width=0.24\linewidth]{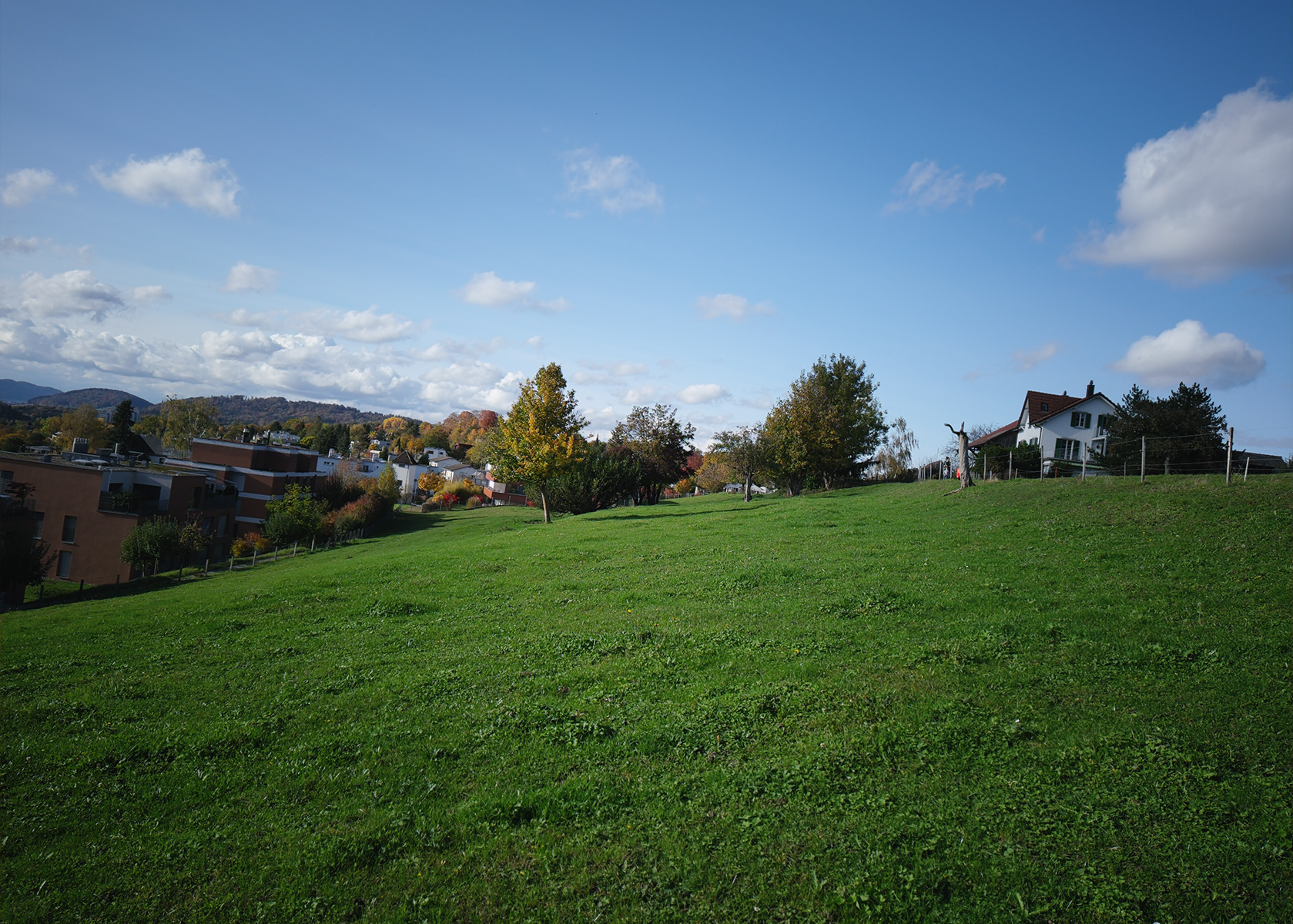}&
    \includegraphics[width=0.24\linewidth]{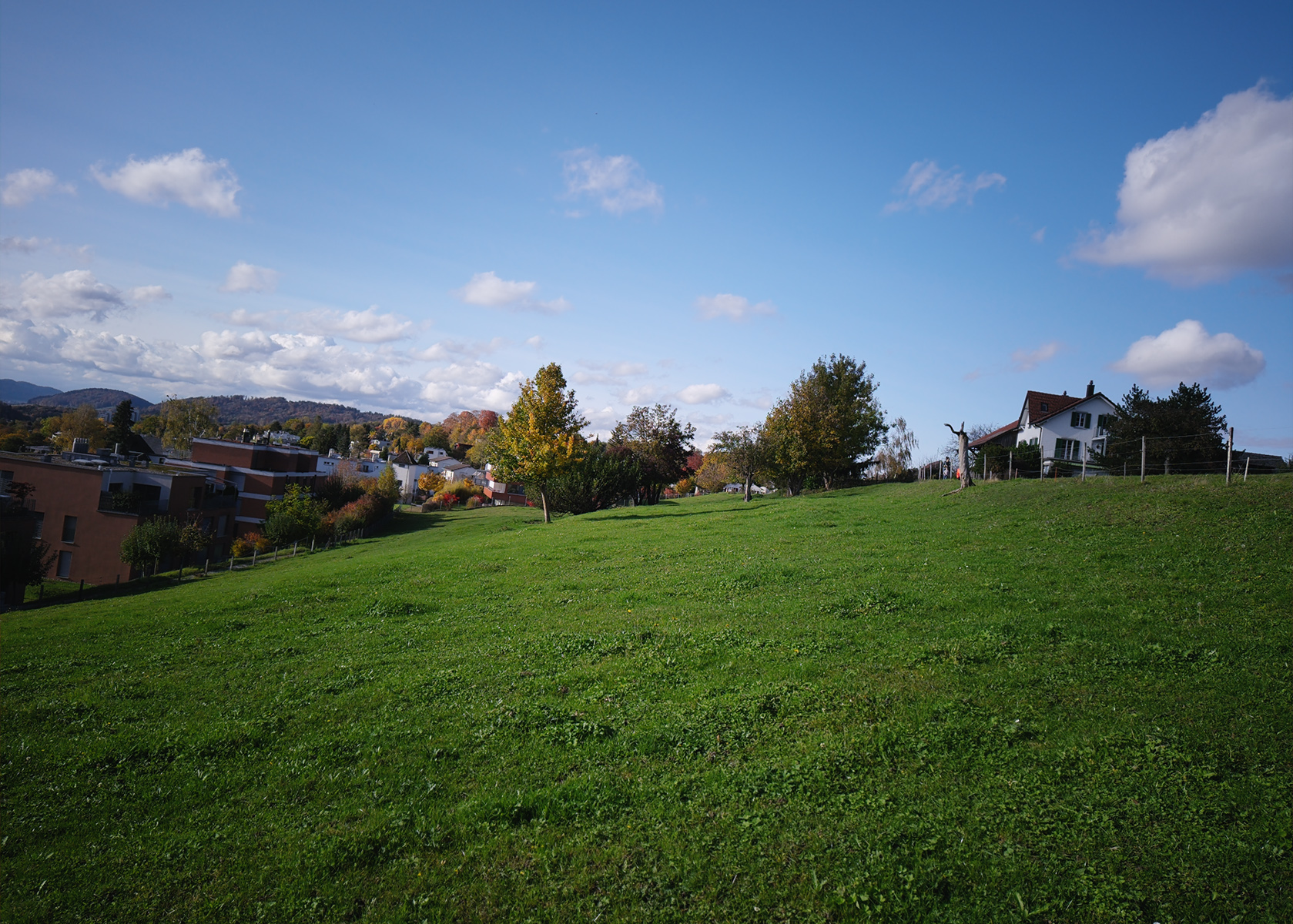}\\
\end{tabular}
}
\vspace{-0.2cm}
\caption{Sample visual results of the MicroISP models with different depth multipliers.}
\label{fig:microisp_depth}
\end{figure*}

\begin{table}[b!]

\centering
\resizebox{0.7\linewidth}{!}
{
\begin{tabular}{l|cc}
\hline
Model Modification \, & \, PSNR \, & \, SSIM \, \\
\hline
\hline
No Attention Block       & 23.25 & 0.835 \\
Standard Attention Block \, & 23.34 & 0.842 \\
\textit{Leaky ReLU} instead of \textit{PReLU} & 23.54 & 0.847 \\
\hline
Final Design              & 23.87 & 0.853 \\
\end{tabular}
}
\vspace{0.8mm}
\caption{The results of the model 1) without the attention block, 2) with the standard attention block using the average pooling, 3) using Leaky ReLU instead of PReLU, 4) the final architecture.}
\label{tab:ablation}
\vspace{-1.2mm}
\end{table}

To evaluate the efficiency of the proposed model design, we performed an ablation study which results are reported in Table~\ref{tab:ablation}. First of all, we checked the importance of the attention unit by either disabling it completely, or by replacing with the conventional implementation utilizing the global average pooling in the first layer. The results demonstrate a very high importance of this building block for the proposed task: thus, the PSNR score drops by more than 0.5dB when this block is removed. While using the standard implementation improves the scores to 23.34 dB, these results are still considerably lower compared to the ones obtained with the proposed attention modification that takes into account image content when computing the normalization coefficients.

Another important design choice was to use the \textit{PReLU} activations instead of the \textit{Leaky ReLU}. While this is almost not affecting the model complexity (as is equivalent to the \textit{Leaky ReLU} with learned slope for 4 input channels), it allows to substantially boost the quality of the reconstructed image results due to using additional global image adjustment parameters. We should also note that the \textit{Leaky ReLU} implementation might still be of interest when running the MicroISP model on older mobile NPUs as they might lack \textit{PReLU} support introduced in Android NNAPI 1.2.

The final principal model parameter is the number of basic building blocks, its effect on the visual and runtime results is discussed in the next section.

\subsection{Adjusting the model complexity}
\label{sec:complexity}

The MicroISP model allows to adapt its computational complexity to the target hardware platform by changing the number of its building units (Fig.~\ref{fig:architecture}). Since the RAM consumption remains almost constant regardless of the model depth, one can potentially design very large or very small networks depending on the task, target runtime and computational budget. Table~\ref{tab:model_complexity} demonstrates the effect of the model size on its fidelity and runtime numbers for the considered ISP problem, while Figure~\ref{fig:microisp_depth} shows the corresponding visual results. As one can see, it is possible to reduce the runtime by more than 33\% by switching to the MicroISP 0.5 network with one building block at the expense of slightly worse image reconstruction quality. If the latency constraints are very tight, one can further reduce the runtime with the MicroISP 0.25 model: even in this case the overall image rendering results are still acceptable, though the texture quality is obviously lower (this, however, might not be very critical \eg, for real-time video processing). In this particular task, any further increase of model complexity is leading to only marginal performance improvements, thus using two building blocks delivers the best runtime-quality trade-off for the considered problem.

\begin{table}[t!]
\centering
\resizebox{\linewidth}{!}
{
\begin{tabular}{l|cc|ccc}
\hline
Model Depth Multiplier \, & \, PSNR \, & \, SSIM \, & \multicolumn{3}{|c}{Runtime on the Dimensity 1000+ GPU} \\
& & & \, FullHD, ms \, &\, 12MP, ms \, & \, 32MP, ms \\
\hline
\hline
MicroISP 1.5\,\,\, [3 blocks] & 23.91 & 0.854 & 56.8 & 314 & 837 \\
MicroISP 1.0\,\,\, [2 blocks] & 23.87 & 0.853 & 42.3 & 238 & 636 \\
MicroISP 0.5\,\,\, [1 block] & 23.60 & 0.846 & 28.2 & 155 & 417 \\
MicroISP 0.25 [half block] & 23.37 & 0.841 & 23.1 & 122 & 336 \\
\end{tabular}
}
\vspace{0.2mm}
\caption{Quantitative and runtime results of several MicroISP models with different depth multipliers.}
\label{tab:model_complexity}
\vspace{-1.2mm}
\end{table}

\subsection{Limitations}

As this is an end-to-end solution, the reconstruction issues on some images are generally inevitable. While the overall quality of the results is high, appr. 5-7\% of images might exhibit an imprecise white balancing, with pinkish or yellowish tones visible in bright photo regions. Next, though the proposed model can handle low and medium noise levels, it cannot suppress heavy noise in images captured at night or in the dark (note however, that the model was not trained for night photo processing as it never observed such image samples). Another global issue is vignetting caused by optics, but this problem can be efficiently fixed with the standard algorithms. Finally, though the experiments revealed that the real resolution of the reconstructed photos is close to the one of the images processed with the classical ISP system, one might want to improve it further as the proposed dataset allows to train the model to perform an additional 2 times image upscaling, though a larger model might be required in this case.

\subsection{PyNET-V2 Mobile}

Besides the MicroISP model, we also developed a considerably more powerful PyNET-V2 Mobile architecture, which structure is inspired by the original PyNET~\cite{ignatov2020replacing} model while its design was fully revised in order to be compatible with mobile AI accelerators. The PyNET-V2 Mobile network achieves a PSNR score of 24.72 dB (+0.85 dB compared to the MicroISP model) on the considered FujiFilm UltraISP dataset, and its runtime on the Dimensity 9000 APU is less than 800 ms when processing raw Full HD resolution images. A detailed description of this architecture and its results can be found in paper~\cite{ignatov2022pynetv2}.

\section{Conclusion}
\label{sec:conclusion}

We proposed a novel deep learning based MicroISP architecture allowing to process RAW photos of resolution up to 32MP directly on mobile devices. The presented solution learns to perform all image processing steps directly from the data, not requiring any manual supervision or hand-crafted features. To check performance, we collected a large-scale Fujifilm UltraISP dataset consisting of more than 6K RAW-RGB image pairs captured by a normal mobile camera sensor and a professional 102MP medium format Fujifilm camera. The experiments revealed that the quality of the reconstructed RGB images is comparable to the results obtained with a classical ISP system, though texture-wise the deep learning based solution is providing superior results. The conducted runtime evaluation shows that the MicroISP model is capable of processing 32MP photos on the majority of mobile SoCs under 500 ms, while for FullHD images it demonstrates real-time performance. Finally, the proposed architecture is also compatible with dedicated mobile AI accelerators such as APUs or NPUs, allowing to further improve the runtime and to reduce power consumption, which might be critical for edge inference.

{\small
\bibliographystyle{ieee_fullname}
\bibliography{egbib}

\begin{thebibliography}{10}\itemsep=-1pt

\bibitem{abdelhamed2020ntire}
Abdelrahman Abdelhamed, Mahmoud Afifi, Radu Timofte, and Michael~S Brown.
\newblock Ntire 2020 challenge on real image denoising: Dataset, methods and
  results.
\newblock In {\em Proceedings of the IEEE/CVF Conference on Computer Vision and
  Pattern Recognition Workshops}, pages 496--497, 2020.

\bibitem{abdelhamed2019ntire}
Abdelrahman Abdelhamed, Radu Timofte, and Michael~S Brown.
\newblock Ntire 2019 challenge on real image denoising: Methods and results.
\newblock In {\em Proceedings of the IEEE/CVF Conference on Computer Vision and
  Pattern Recognition Workshops}, pages 0--0, 2019.

\bibitem{NNAPIDrivers2021}
Android Neural~Networks API.
\newblock https://source.android.com/devices/neural-networks.

\bibitem{cai2019ntire}
Jianrui Cai, Shuhang Gu, Radu Timofte, and Lei Zhang.
\newblock Ntire 2019 challenge on real image super-resolution: Methods and
  results.
\newblock In {\em Proceedings of the IEEE/CVF Conference on Computer Vision and
  Pattern Recognition Workshops}, pages 0--0, 2019.

\bibitem{cai2018learning}
Jianrui Cai, Shuhang Gu, and Lei Zhang.
\newblock Learning a deep single image contrast enhancer from multi-exposure
  images.
\newblock {\em IEEE Transactions on Image Processing}, 27(4):2049--2062, 2018.

\bibitem{dai2020awnet}
Linhui Dai, Xiaohong Liu, Chengqi Li, and Jun Chen.
\newblock Awnet: Attentive wavelet network for image isp.
\newblock {\em arXiv preprint arXiv:2008.09228}, 2020.

\bibitem{de2018fast}
Etienne de Stoutz, Andrey Ignatov, Nikolay Kobyshev, Radu Timofte, and Luc
  Van~Gool.
\newblock Fast perceptual image enhancement.
\newblock In {\em Proceedings of the European Conference on Computer Vision
  (ECCV) Workshops}, pages 0--0, 2018.

\bibitem{dong2015image}
Chao Dong, Chen~Change Loy, Kaiming He, and Xiaoou Tang.
\newblock Image super-resolution using deep convolutional networks.
\newblock {\em IEEE transactions on pattern analysis and machine intelligence},
  38(2):295--307, 2015.

\bibitem{dong2016accelerating}
Chao Dong, Chen~Change Loy, and Xiaoou Tang.
\newblock Accelerating the super-resolution convolutional neural network.
\newblock In {\em European conference on computer vision}, pages 391--407.
  Springer, 2016.

\bibitem{fu2016fusion}
Xueyang Fu, Delu Zeng, Yue Huang, Yinghao Liao, Xinghao Ding, and John Paisley.
\newblock A fusion-based enhancing method for weakly illuminated images.
\newblock {\em Signal Processing}, 129:82--96, 2016.

\bibitem{gu2019brief}
Shuhang Gu and Radu Timofte.
\newblock A brief review of image denoising algorithms and beyond.
\newblock {\em Inpainting and Denoising Challenges}, pages 1--21, 2019.

\bibitem{hsyu2021CSANet}
Ming-Chun Hsyu, Chih-Wei Liu, Chao-Hung Chen, Chao-Wei Chen, and Wen-Chia Tsai.
\newblock Csanet: High speed channel spatial attention network for mobile isp.
\newblock In {\em Proceedings of the IEEE/CVF Conference on Computer Vision and
  Pattern Recognition Workshops}, pages 0--0, 2021.

\bibitem{huang2018range}
Jie Huang, Pengfei Zhu, Mingrui Geng, Jiewen Ran, Xingguang Zhou, Chen Xing,
  Pengfei Wan, and Xiangyang Ji.
\newblock Range scaling global u-net for perceptual image enhancement on mobile
  devices.
\newblock In {\em Proceedings of the European Conference on Computer Vision
  (ECCV) Workshops}, pages 0--0, 2018.

\bibitem{hui2018perception}
Zheng Hui, Xiumei Wang, Lirui Deng, and Xinbo Gao.
\newblock Perception-preserving convolutional networks for image enhancement on
  smartphones.
\newblock In {\em Proceedings of the European Conference on Computer Vision
  (ECCV) Workshops}, pages 0--0, 2018.

\bibitem{ignatov2021fast}
Andrey Ignatov, Kim Byeoung-su, Radu Timofte, and Angeline Pouget.
\newblock Fast camera image denoising on mobile gpus with deep learning, mobile
  ai 2021 challenge: Report.
\newblock In {\em Proceedings of the IEEE/CVF Conference on Computer Vision and
  Pattern Recognition}, pages 2515--2524, 2021.

\bibitem{ignatov2021learned}
Andrey Ignatov, Jimmy Chiang, Hsien-Kai Kuo, Anastasia Sycheva, and Radu
  Timofte.
\newblock Learned smartphone isp on mobile npus with deep learning, mobile ai
  2021 challenge: Report.
\newblock In {\em Proceedings of the IEEE/CVF Conference on Computer Vision and
  Pattern Recognition Workshops}, pages 0--0, 2021.

\bibitem{ignatov2017dslr}
Andrey Ignatov, Nikolay Kobyshev, Radu Timofte, Kenneth Vanhoey, and Luc
  Van~Gool.
\newblock Dslr-quality photos on mobile devices with deep convolutional
  networks.
\newblock In {\em Proceedings of the IEEE International Conference on Computer
  Vision}, pages 3277--3285, 2017.

\bibitem{ignatov2022pynetv2}
Andrey Ignatov, Grigory Malivenko, Radu Timofte, Yu Tseng, Yu-Syuan Xu,
  Po-Hsiang Yu, Cheng-Ming Chiang, Hsien-Kai Kuo, Min-Hung Chen, Chia-Ming
  Cheng, and Luc Van~Gool.
\newblock Pynet-v2 mobile: Efficient on-device photo processing with neural
  networks.
\newblock In {\em 2021 26th International Conference on Pattern Recognition
  (ICPR)}. IEEE, 2022.

\bibitem{ignatov2019ntire}
Andrey Ignatov and Radu Timofte.
\newblock Ntire 2019 challenge on image enhancement: Methods and results.
\newblock In {\em Proceedings of the IEEE/CVF Conference on Computer Vision and
  Pattern Recognition Workshops}, pages 0--0, 2019.

\bibitem{ignatov2018ai}
Andrey Ignatov, Radu Timofte, William Chou, Ke Wang, Max Wu, Tim Hartley, and
  Luc Van~Gool.
\newblock Ai benchmark: Running deep neural networks on android smartphones.
\newblock In {\em Proceedings of the European Conference on Computer Vision
  (ECCV) Workshops}, pages 0--0, 2018.

\bibitem{ignatov2021real}
Andrey Ignatov, Radu Timofte, Maurizio Denna, and Abdel Younes.
\newblock Real-time quantized image super-resolution on mobile npus, mobile ai
  2021 challenge: Report.
\newblock In {\em Proceedings of the IEEE/CVF Conference on Computer Vision and
  Pattern Recognition Workshops}, pages 0--0, 2021.

\bibitem{ignatov2019aim}
Andrey Ignatov, Radu Timofte, Sung-Jea Ko, Seung-Wook Kim, Kwang-Hyun Uhm,
  Seo-Won Ji, Sung-Jin Cho, Jun-Pyo Hong, Kangfu Mei, Juncheng Li, et~al.
\newblock Aim 2019 challenge on raw to rgb mapping: Methods and results.
\newblock In {\em 2019 IEEE/CVF International Conference on Computer Vision
  Workshop (ICCVW)}, pages 3584--3590. IEEE, 2019.

\bibitem{ignatov2019ai}
Andrey Ignatov, Radu Timofte, Andrei Kulik, Seungsoo Yang, Ke Wang, Felix Baum,
  Max Wu, Lirong Xu, and Luc Van~Gool.
\newblock Ai benchmark: All about deep learning on smartphones in 2019.
\newblock In {\em 2019 IEEE/CVF International Conference on Computer Vision
  Workshop (ICCVW)}, pages 3617--3635. IEEE, 2019.

\bibitem{ignatov2018pirm}
Andrey Ignatov, Radu Timofte, Thang Van~Vu, Tung Minh~Luu, Trung X~Pham, Cao
  Van~Nguyen, Yongwoo Kim, Jae-Seok Choi, Munchurl Kim, Jie Huang, et~al.
\newblock Pirm challenge on perceptual image enhancement on smartphones:
  Report.
\newblock In {\em Proceedings of the European Conference on Computer Vision
  (ECCV) Workshops}, pages 0--0, 2018.

\bibitem{ignatov2020aim}
Andrey Ignatov, Radu Timofte, Zhilu Zhang, Ming Liu, Haolin Wang, Wangmeng Zuo,
  Jiawei Zhang, Ruimao Zhang, Zhanglin Peng, Sijie Ren, et~al.
\newblock Aim 2020 challenge on learned image signal processing pipeline.
\newblock {\em arXiv preprint arXiv:2011.04994}, 2020.

\bibitem{ignatov2020replacing}
Andrey Ignatov, Luc Van~Gool, and Radu Timofte.
\newblock Replacing mobile camera isp with a single deep learning model.
\newblock In {\em Proceedings of the IEEE/CVF Conference on Computer Vision and
  Pattern Recognition Workshops}, pages 536--537, 2020.

\bibitem{johnson2016perceptual}
Justin Johnson, Alexandre Alahi, and Li Fei-Fei.
\newblock Perceptual losses for real-time style transfer and super-resolution.
\newblock In {\em European conference on computer vision}, pages 694--711.
  Springer, 2016.

\bibitem{kim2020pynet}
Byung-Hoon Kim, Joonyoung Song, Jong~Chul Ye, and JaeHyun Baek.
\newblock Pynet-ca: enhanced pynet with channel attention for end-to-end mobile
  image signal processing.
\newblock In {\em European Conference on Computer Vision}, pages 202--212.
  Springer, 2020.

\bibitem{kim2016accurate}
Jiwon Kim, Jung~Kwon Lee, and Kyoung~Mu Lee.
\newblock Accurate image super-resolution using very deep convolutional
  networks.
\newblock In {\em Proceedings of the IEEE conference on computer vision and
  pattern recognition}, pages 1646--1654, 2016.

\bibitem{kingma2014adam}
Diederik~P Kingma and Jimmy Ba.
\newblock Adam: A method for stochastic optimization.
\newblock {\em arXiv preprint arXiv:1412.6980}, 2014.

\bibitem{lee2019device}
Juhyun Lee, Nikolay Chirkov, Ekaterina Ignasheva, Yury Pisarchyk, Mogan Shieh,
  Fabio Riccardi, Raman Sarokin, Andrei Kulik, and Matthias Grundmann.
\newblock On-device neural net inference with mobile gpus.
\newblock {\em arXiv preprint arXiv:1907.01989}, 2019.

\bibitem{lim2017enhanced}
Bee Lim, Sanghyun Son, Heewon Kim, Seungjun Nah, and Kyoung Mu~Lee.
\newblock Enhanced deep residual networks for single image super-resolution.
\newblock In {\em Proceedings of the IEEE conference on computer vision and
  pattern recognition workshops}, pages 136--144, 2017.

\bibitem{liu2018deep}
Hanwen Liu, Pablo Navarrete~Michelini, and Dan Zhu.
\newblock Deep networks for image-to-image translation with mux and demux
  layers.
\newblock In {\em Proceedings of the European Conference on Computer Vision
  (ECCV) Workshops}, pages 0--0, 2018.

\bibitem{lugmayr2019unsupervised}
Andreas Lugmayr, Martin Danelljan, and Radu Timofte.
\newblock Unsupervised learning for real-world super-resolution.
\newblock In {\em 2019 IEEE/CVF International Conference on Computer Vision
  Workshop (ICCVW)}, pages 3408--3416. IEEE, 2019.

\bibitem{lugmayr2020ntire}
Andreas Lugmayr, Martin Danelljan, and Radu Timofte.
\newblock Ntire 2020 challenge on real-world image super-resolution: Methods
  and results.
\newblock In {\em Proceedings of the IEEE/CVF Conference on Computer Vision and
  Pattern Recognition Workshops}, pages 494--495, 2020.

\bibitem{ma2014high}
Kede Ma, Hojatollah Yeganeh, Kai Zeng, and Zhou Wang.
\newblock High dynamic range image tone mapping by optimizing tone mapped image
  quality index.
\newblock In {\em 2014 IEEE International Conference on Multimedia and Expo
  (ICME)}, pages 1--6. IEEE, 2014.

\bibitem{ronneberger2015u}
Olaf Ronneberger, Philipp Fischer, and Thomas Brox.
\newblock U-net: Convolutional networks for biomedical image segmentation.
\newblock In {\em International Conference on Medical image computing and
  computer-assisted intervention}, pages 234--241. Springer, 2015.

\bibitem{salih2012tone}
Yasir Salih, Aamir~S Malik, Naufal Saad, et~al.
\newblock Tone mapping of hdr images: A review.
\newblock In {\em 2012 4th International Conference on Intelligent and Advanced
  Systems (ICIAS2012)}, volume~1, pages 368--373. IEEE, 2012.

\bibitem{silva2020deep}
Jose Ivson~S Silva, Gabriel~G Carvalho, Marcel~Santana Santos, Diego~JC
  Santiago, Lucas~Pontes de Albuquerque, Jorge F~Puig Battle, Gabriel~M da
  Costa, and Tsang~Ing Ren.
\newblock A deep learning approach to mobile camera image signal processing.
\newblock In {\em Anais Estendidos do XXXIII Conference on Graphics, Patterns
  and Images}, pages 225--231. SBC, 2020.

\bibitem{NNAPI10Specs}
Android Neural Networks API~1.0 Specifications.
\newblock https://android.googlesource.com/platform/hardware/interfaces/
  +/refs/heads/master/neuralnetworks/1.0/types.hal.

\bibitem{NNAPI12Specs}
Android Neural Networks API~1.2 Specifications.
\newblock https://android.googlesource.com/platform/hardware/interfaces/
  +/refs/heads/master/neuralnetworks/1.2/types.hal.

\bibitem{NNAPI13Specs}
Android Neural Networks API~1.3 Specifications.
\newblock https://android.googlesource.com/platform/hardware/interfaces/
  +/refs/heads/master/neuralnetworks/1.3/types.hal.

\bibitem{tai2017memnet}
Ying Tai, Jian Yang, Xiaoming Liu, and Chunyan Xu.
\newblock Memnet: A persistent memory network for image restoration.
\newblock In {\em Proceedings of the IEEE international conference on computer
  vision}, pages 4539--4547, 2017.

\bibitem{timofte2017ntire}
Radu Timofte, Eirikur Agustsson, Luc Van~Gool, Ming-Hsuan Yang, and Lei Zhang.
\newblock Ntire 2017 challenge on single image super-resolution: Methods and
  results.
\newblock In {\em Proceedings of the IEEE conference on computer vision and
  pattern recognition workshops}, pages 114--125, 2017.

\bibitem{timofte2018ntire}
Radu Timofte, Shuhang Gu, Jiqing Wu, and Luc Van~Gool.
\newblock Ntire 2018 challenge on single image super-resolution: Methods and
  results.
\newblock In {\em Proceedings of the IEEE conference on computer vision and
  pattern recognition workshops}, pages 852--863, 2018.

\bibitem{truong2021learning}
Prune Truong, Martin Danelljan, Luc Van~Gool, and Radu Timofte.
\newblock Learning accurate dense correspondences and when to trust them.
\newblock {\em arXiv preprint arXiv:2101.01710}, 2021.

\bibitem{vu2018fast}
Thang Vu, Cao Van~Nguyen, Trung~X Pham, Tung~M Luu, and Chang~D Yoo.
\newblock Fast and efficient image quality enhancement via desubpixel
  convolutional neural networks.
\newblock In {\em Proceedings of the European Conference on Computer Vision
  (ECCV) Workshops}, pages 0--0, 2018.

\bibitem{wang2018esrgan}
Xintao Wang, Ke Yu, Shixiang Wu, Jinjin Gu, Yihao Liu, Chao Dong, Yu Qiao, and
  Chen Change~Loy.
\newblock Esrgan: Enhanced super-resolution generative adversarial networks.
\newblock In {\em Proceedings of the European Conference on Computer Vision
  (ECCV) Workshops}, pages 0--0, 2018.

\bibitem{yan2016automatic}
Zhicheng Yan, Hao Zhang, Baoyuan Wang, Sylvain Paris, and Yizhou Yu.
\newblock Automatic photo adjustment using deep neural networks.
\newblock {\em ACM Transactions on Graphics (TOG)}, 35(2):11, 2016.

\bibitem{lee2016automatic}
Zhicheng Yan, Hao Zhang, Baoyuan Wang, Sylvain Paris, and Yizhou Yu.
\newblock Automatic photo adjustment using deep neural networks.
\newblock volume~35, page~11. ACM, 2016.

\bibitem{yuan2012automatic}
Lu Yuan and Jian Sun.
\newblock Automatic exposure correction of consumer photographs.
\newblock In {\em European Conference on Computer Vision}, pages 771--785.
  Springer, 2012.

\bibitem{zhang2020ntire}
Kai Zhang, Shuhang Gu, and Radu Timofte.
\newblock Ntire 2020 challenge on perceptual extreme super-resolution: Methods
  and results.
\newblock In {\em Proceedings of the IEEE/CVF Conference on Computer Vision and
  Pattern Recognition Workshops}, pages 492--493, 2020.

\bibitem{zhang2017beyond}
Kai Zhang, Wangmeng Zuo, Yunjin Chen, Deyu Meng, and Lei Zhang.
\newblock Beyond a gaussian denoiser: Residual learning of deep cnn for image
  denoising.
\newblock {\em IEEE transactions on image processing}, 26(7):3142--3155, 2017.

\bibitem{zhang2018ffdnet}
Kai Zhang, Wangmeng Zuo, and Lei Zhang.
\newblock Ffdnet: Toward a fast and flexible solution for cnn-based image
  denoising.
\newblock {\em IEEE Transactions on Image Processing}, 27(9):4608--4622, 2018.

\end{thebibliography}
}

\end{document}